\documentclass[referee,pdflatex,sn-mathphys-num,12pt]{sn-jnl}%

\usepackage[T1]{fontenc}
\usepackage[utf8]{inputenc}
\usepackage{libertine}

\usepackage{graphicx}%
\usepackage{amsmath,amssymb,amsfonts}%
\usepackage{amsthm}%
\usepackage{mathrsfs}%
\usepackage{xcolor}%
\usepackage{textcomp}%
\usepackage{manyfoot}%
\usepackage{booktabs}%
\usepackage{algorithm}%
\usepackage{algorithmicx}%
\usepackage{algpseudocode}%
\usepackage{listings}%
\usepackage{multicol}
\usepackage{array}
\usepackage{subcaption}
\usepackage{hyperref}
\hypersetup{
    colorlinks = true,
    linkcolor = blue,
    urlcolor= blue,
    pdftitle={Teleoperation System},}

\begin{document}

\title{\LARGE \bf A Master-Follower Teleoperation System for Robotic Catheterization: Design, Characterization, and Tracking Control}

%%=============================================================%%
%%=============================================================%%

\author*[1]{\fnm{Ali A.} \sur{Nazari}}\email{ali.nazari@torontomu.ca}

\author[1]{\fnm{Jeremy} \sur{Catania}}%\email{jcatania@torontomu.ca}

\author[2]{\fnm{Soroush} \sur{Sadeghian}}%\email{soroush.sadeghian@torontomu.ca}

\author[1]{\fnm{Amir} \sur{Jalali}}%\email{amirjalali58@gmail.com}

\author[1]{\fnm{Houman} \sur{Masnavi}}%\email{houman@torontomu.ca}

\author[1]{\fnm{Farrokh} \sur{Janabi-Sharifi}}%\email{fsharifi@torontomu.ca}

\author[1]{\fnm{Kourosh} \sur{Zareinia}}%\email{kourosh.zareinia@torontomu.ca}

\affil[1]{\orgdiv{Department of Mechanical, Industrial and Mechatronics Engineering}, \orgname{Toronto Metropolitan University}, \orgaddress{\street{350 Victoria St.}, \city{Toronto}, \postcode{M5B 2K3}, \state{ON}, \country{Canada}}}

\affil[2]{\orgdiv{Department of Computer Science}, \orgname{Toronto Metropolitan University}, \orgaddress{\street{350 Victoria St.}, \city{Toronto}, \postcode{M5B 2K3}, \state{ON}, \country{Canada}}}

%%==================================%%
%% Abstract %%
%%==================================%%

\abstract{
\textbf{Background:} Over the past two decades, telerobotic systems with robot-mediated, minimally invasive techniques, have mitigated radiation exposure for medical staff and extended medical services to remote areas. To enhance these services, master-follower telerobotic systems should offer transparency, enabling surgeons and clinicians to feel force interactions similar to those the follower device experiences with patients' bodies.

\textbf{Methods:} We present a three-degree-of-freedom master-follower teleoperated system for robotic catheterization. The follower device uses a grip-insert-release mechanism to prevent catheter buckling and torsion, mimicking real-world manual intervention. Performance is evaluated through open-loop path tracking on circular, infinity-like, and spiral paths.

\textbf{Results:} Path tracking errors, mean Euclidean error (MEE) and mean absolute error (MAE), range from 0.64 to 1.53 cm (MEE) and 0.81 to 1.92 cm (MAE) for different paths.

\textbf{Conclusion:} While the system meets precision and accuracy targets with an open-loop controller, closed-loop control is needed to address catheter hysteresis, dead zones, and nonlinearities.
} 

\keywords{Minimally invasive surgery, Robotic catheterization, Master-follower teleoperation system, Characterization, Tracking control}

\maketitle

\section{Introduction} \label{sec:intro}

Technological advancements in robotics have led to effective solutions for complex medical interventions in recent decades \cite{dupont2021decade,simaan2018medical,ginoya2021historical,chen2023catheter,ren2023critical,zhao2022remote,alekseeva2024design,li2024robotic}. Robotic systems, with their high accuracy and dexterity, have facilitated minimally invasive surgery (MIS) and various medical procedures, including heart intervention. These systems can reduce the side effects of invasive surgeries, such as post-surgical trauma and hospital stays. 

Catheterization, a common heart intervention, stands to benefit significantly from advanced robotic technologies \cite{rafii2014current}. Catheters, used in procedures like cardiovascular \cite{tavallaei2016design,lee2018mr,choi2021vascular,baek2022design,rosch2023enhanced,qi2023telerobotic,yu2023design,peng2023design}, endovascular \cite{he2018linear,abdelaziz2019toward,guo2019novel,kundrat2021mr,li2022endovascular,alekseeva2024design,li2024robotic}, neurovascular\cite{kim2022telerobotic,wei2024telemanipulated}, and urological interventions \cite{franco2023robot}, can be viewed as medical continuum robots due to their flexible, shape-deformable structures. In heart interventions, catheter tip movement is typically controlled using tendon-driven mechanisms, such as strings and knobs \cite{nazari2022visual}.

Cardiac catheterization procedures present significant challenges in modern healthcare. While these minimally invasive techniques have revolutionized cardiovascular care, they expose medical staff to considerable cumulative radiation doses during their careers, particularly affecting interventional cardiologists and support staff who perform procedures regularly \cite{kobayashi2017radiation}. This occupational hazard is compounded by the growing volume and complexity of catheterization procedures worldwide. Moreover, these interventions demand sustained precision over extended periods, often lasting several hours, leading to operator fatigue and potential complications. Considering the costs of radiation protection equipment per catheterization site and the impact of ergonomic injuries among practitioners, the economic burden is substantial. This burden can lead to reduced productivity and early career termination. These challenges, combined with the growing complexity of interventional procedures and the need to extend specialized cardiac care to underserved regions, underscore the necessity for innovative robotic solutions that can enhance both operator safety and procedural outcomes.

Robotic catheterization offers numerous advantages for clinicians, enhancing efficiency in repetitive and lengthy procedures \cite{rafii2014current}. However, the issue of radiation exposure (from X-ray, fluoroscopy, and CT scan) remains a significant concern. While protective solutions exist in medical sites, they often result in clinician fatigue, underscoring the need for more effective safety mechanisms.

To address this issue, telerobotic systems have emerged in the last two decades to place operators at safe workstations located in radiation-free zones \cite{chen2023catheter}. These systems employ a master device controlled by clinicians and a follower device in contact with the patient. This setup protects operators from radiation exposure.

Effective telerobotic interventions depend on reliable communication channels between the master and follower devices, spanning distances from a few meters to several thousand kilometers \cite{farajiparvar2020brief}. Real-time, bilateral feedback is essential for providing clinicians with immediate force feedback, ensuring that the follower device accurately replicates the master device's movements \cite{shahbazi2018systematic}. This high-fidelity replication allows clinicians to operate as if they were directly manipulating the catheter, thereby improving precision and patient outcomes. Designing these systems requires adherence to rigorous medical engineering standards to ensure both safety and efficacy in clinical settings \cite{torabi2021kinematic,patel2022haptic}. 

Building on these foundations and requirements, various telerobotic catheterization systems have been developed, each attempting to address specific clinical needs and technical challenges. The following section reviews these developments and identifies remaining opportunities for improvement.

\subsection{Related Works} \label{subsec:prior}

In recent years, various telerobotic systems have been introduced to enhance the performance of medical interventions \cite{zhao2022remote}. Huang \textit{et al.} \cite{huang2021three} presented a three-limb, 13-degree-of-freedom (DOF) telerobotic endoscopic system controlled, continuously and simultaneously, using one foot and two hands. The system included two omega.7 (Force Dimension, Switzerland) haptic interfaces in the master device, and one grasper and one cauterizing hook in the follower device. However, the follower's bulky design posed occupancy issues in operating rooms. 

Yu \textit{et al.} \cite{yu2023design} developed a vascular interventional robot capable of delivering multiple instruments remotely. While the 15-motor system ensured accurate actuation, it required a complex controller. The follower device's roller-type catheter navigation mechanism was also prone to torsional effects \cite{yu2023design}. Bao \textit{et al.} \cite{bao2019design} introduced a master-follower system for vascular surgeries, but could not fully simulate realistic medical interventions due to the use of two commercial master devices with functionalities different from catheters/guidewires. 

Choi \textit{et al.} \cite{choi2021vascular} proposed a 3-DOF modularized teleoperation system for cardiovascular interventions with a novel master console design consisting of planar and spherical mechanisms and a linear actuator. The follower device also included three roller cartridge-based modules for the delivery of interventional tools, but the system did not mimic the real-world catheter handle, increasing cognitive load for clinicians. Tavallaei \textit{et al.} \cite{tavallaei2016design} designed a 3-DOF telerobotic system for remote catheter navigation and evaluated its performance in \textit{ex-vivo} and \textit{in-vivo} ablation, but lacked haptic feedback for interventionalists. 

Kundrat \textit{et al.} \cite{kundrat2021mr} proposed a pneumatically-actuated magnetic resonance (MR)-safe master-follower system with 2-DOF master and 4-DOF follower devices to manipulate off-the-shelf angiographic catheters and guidewires. The follower device included four platforms for catheter/guidewire feeding and following/rotation \cite{abdelaziz2019toward}. They tested the system under fluoroscopic guidance and haptic feedback on abdominal and thoracic phantoms. Despite mimicing conventional manual catheter manipulation, catheter buckling occurred in unsuccessful trials due to friction between the catheter, the introducer, and the vascular phantom.

Takagi \textit{et al.} \cite{takagi2023development} developed a 2-DOF master-follower system with high-speed force, but without tip bending control. Peng \textit{et al.} \cite{peng2023design} proposed a dual-use mechanism for robot-assisted cardiovascular intervention, incorporating a roller-type mechanism. The mechanism functioned as both master and follower devices, enhancing transparency in teleoperation. The system, however, lacked haptic feedback, preventing interventionalists from manipulating the intervention tool directly.

Furthermore, several other reports have introduced designs for either master or follower devices, tailored for specific tasks with required DOFs and complex mechanical designs \cite{yu2024design,qi2023telerobotic,takagi2023development,francis2018design,he2018linear}. These include commercially available devices \cite{moon2018novel}, hydraulically actuated MR-safe systems for cardiac electrophysiological intervention \cite{lee2018mr}, and magnetorheological fluid-based master devices for endovascular catheterization \cite{guo2019novel}.

In our previous research, we introduced Althea II \cite{norouzi2021design}, a 3-DOF robotic system designed for robotic cardiovascular interventions. Although the system featured twist, translation, and bending capabilities as a potential follower device in teleoperated catheterization, its angled structure increased the likelihood of catheter buckling despite its portability and adaptability to different catheters. Building on these insights, we present a new 3-DOF telerobotic system featuring a grip-insert-release mechanism that closely resembles a conventional catheter handle. This mechanism represents a significant improvement over conventional roller-type designs by mimicking natural manipulation techniques used by clinicians, reducing both catheter stress and operator learning curves. The difference between the roller-type and the new grip-insert-release designs will be illustrated through figures in the next section. The cost-effective modular construction of the new design maintains a compact footprint suitable for space-constrained cardiac catheterization sites. The system provides force feedback capabilities for bilateral teleoperation, shared control in medical interventions, and delicate inspections in confined spaces. The master device was previously evaluated in a simulated environment to assess surgical safety and navigation within confined spaces of the human body \cite{abbasi2024haptic}. The virtual fixture technique from that study can be integrated into this new system to enhance safety and task performance in real surgical scenarios.

The proposed master-follower teleoperation system targets cardiac catheterization procedures, particularly cardiac ablation and complex cardiovascular interventions. The system is specifically designed to handle the complex bidirectional bending requirements of cardiac ablation catheters, which can require significant forces for tip deflection and precise control to reach various anatomical targets. While autonomous robotic systems have shown promise, teleoperation offers distinct advantages for these procedures. First, cardiac catheterization requires real-time decision-making based on complex anatomical structures, patient-specific conditions, and dynamic physiological responses; thus, direct clinician control and procedural experience are crucial. Second, cardiovascular interventions often involve unexpected situations such as anatomical variations, vessel spasms, or arrhythmias that require immediate adaptation, which current autonomous systems cannot handle adequately and flexibly \cite{koskinopoulou2023robotic}. Third, teleoperated systems offer a more direct regulatory pathway for high-risk procedures while bridging the gap between manual techniques and future autonomous interventions through enhanced precision and reduced radiation exposure \cite{attanasio2021autonomy,haidegger2022robot}.

The contributions of this paper are:
\begin{itemize}
    \item Proposing a 3-DOF haptic-enabled teleoperated system for robotic catheterization, with a novel grip-insert-release mechanism in the follower device which resembles the manual operation of clinicians and eliminates catheter torsion and buckling issues during operation;
    \item Characterizing a bidirectionally navigable ablation catheter, for the first time, for teleoperated interaction control in realistic medical experiments;
    \item Validating the telerobotic system's functionality through open-loop approaching and path-tracking experiments.
\end{itemize}

%%%%%%%%%%%%%%%%%%%%%%%%%%%%%%%%%%%%%%%%%%%%%%%%%%%%%%%%%%%%%%%
\section{Materials and Methods} \label{sec:material&methods}

In this research, none of the authors performed any studies with human participants or animals. In the following sections, we will provide more information about different stages of the research.

\subsection{System Design} \label{subsec:design}

The proposed teleoperation system includes 3-DOF master and follower devices. The follower device consists of two parts: the handler and feeder mechanisms. The following subsections elaborate on these components.

\subsubsection{Master Device} \label{subsubsec:masterDesign}

\begin{figure*}
    \begin{center}
        \includegraphics[width=\textwidth]{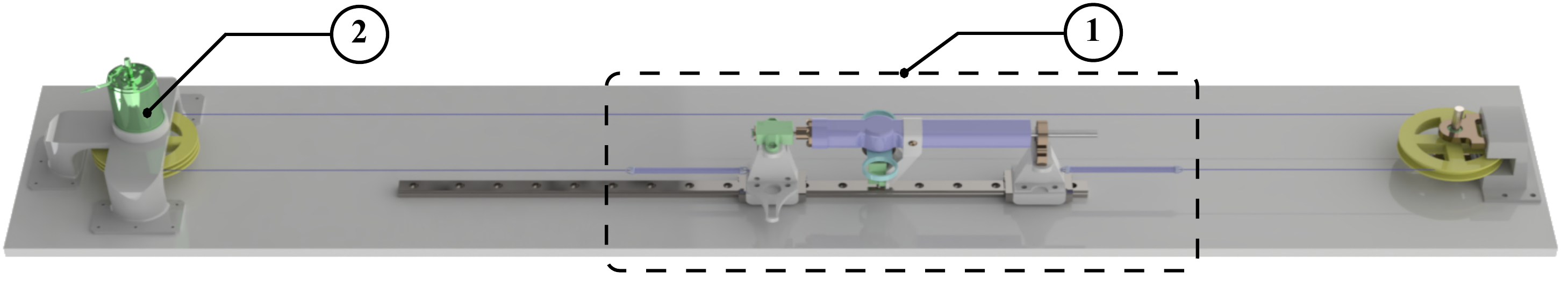}
        \caption{3D CAD model of the master device: (1) master device's handle; (2) translational reaction motor.}
        \label{fig:masterCAD}
        \includegraphics[width=0.85\textwidth]{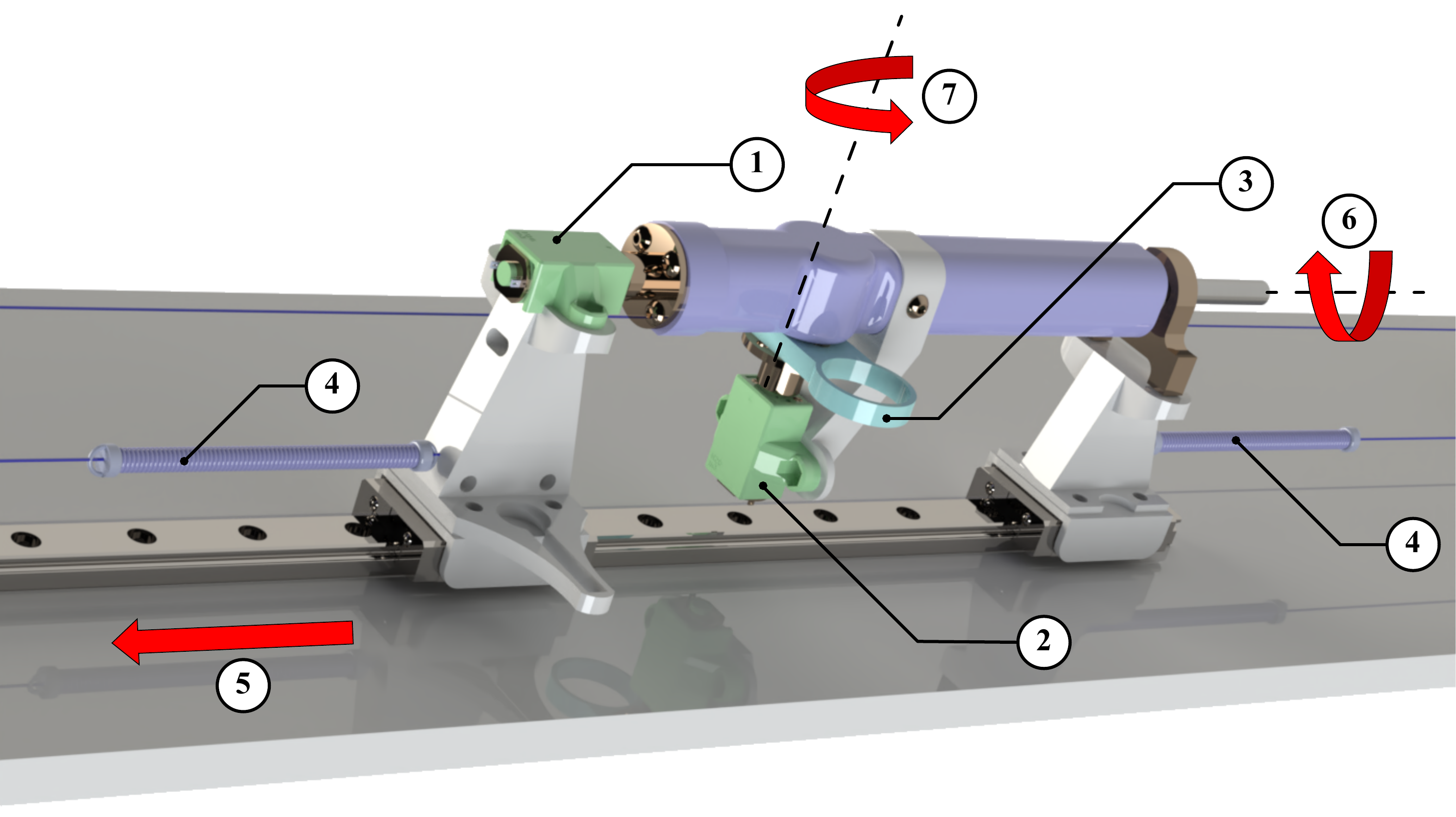}
        \caption{3D CAD model of the master device's handle: (1) rotational reaction motor; (2) bending reaction motor; (3) bending knob; (4) linear spring; (5) direction of translation; (6) axis of rotation - catheter handle; (7) axis of rotation - bending knob/motor.}
        \label{fig:masterHandle}
    \end{center}
\end{figure*}

The master device models three DOFs of a tendon-driven ablation catheter, comprising the shaft, handle, and knob. As shown in Figs. \ref{fig:masterCAD} and \ref{fig:masterHandle}, the master device represents the handle assembly that controls the catheter's translation, rotation, and bending. The catheter tip is bent using the knob. 

Three DC motors apply reaction forces in these DOFs, allowing the operators to feel realistic resistance in their fingers, wrist, and arm while actuating the device's bending, rotation, and translation respectively. The motors are sized to match the feedback needed for each body part of the human operator. For instance, the motor with the smallest torque provides bending feedback to the operator's fingers via the knob, while the motor with the largest torque offers translational feedback to the operator's arm through the handle.

Two DC motors are mounted on the handle to actuate bending and rotational reaction forces, while a third DC motor, mounted on the base, drives translational reaction forces via a pulley system. The handle's axis aligns with the rotational motor shaft, whereas the bending motor is oriented perpendicular to the handle. The bending motor's center of mass is offset from the handle's axis, resulting in an undesired inertial effect on the catheter handle. To mitigate this unwanted torque, the rotational motor is equipped with a higher gear ratio, thereby providing greater torque compared to the bending motor. Otherwise, both DC motors are identical in design and specification.

The translational motor, mounted on the base, minimizes adverse inertial effects on the system dynamics. The handle assembly, which includes the knob and two motors, travels along a linear rail, reducing friction and restricting motion to the longitudinal axis. Due to the relatively large mass of the handle assembly, the translational motor is more powerful than the other two DC motors. A dual pulley system and linear rail convert the translational motor's motion to linear translation. As depicted in Fig. \ref{fig:masterHandle}, pulleys are mounted onto the base at both ends of the linear rail. These pulleys are connected to each other via the handle with a cord, the cord's end is attached to the handle using a linear spring to maintain continuous tension in the cord.

\subsubsection{Follower Device} \label{subsubsec:followerDesign}

\begin{figure*}%[thpb]
    \begin{center}
        \includegraphics[width=\textwidth]{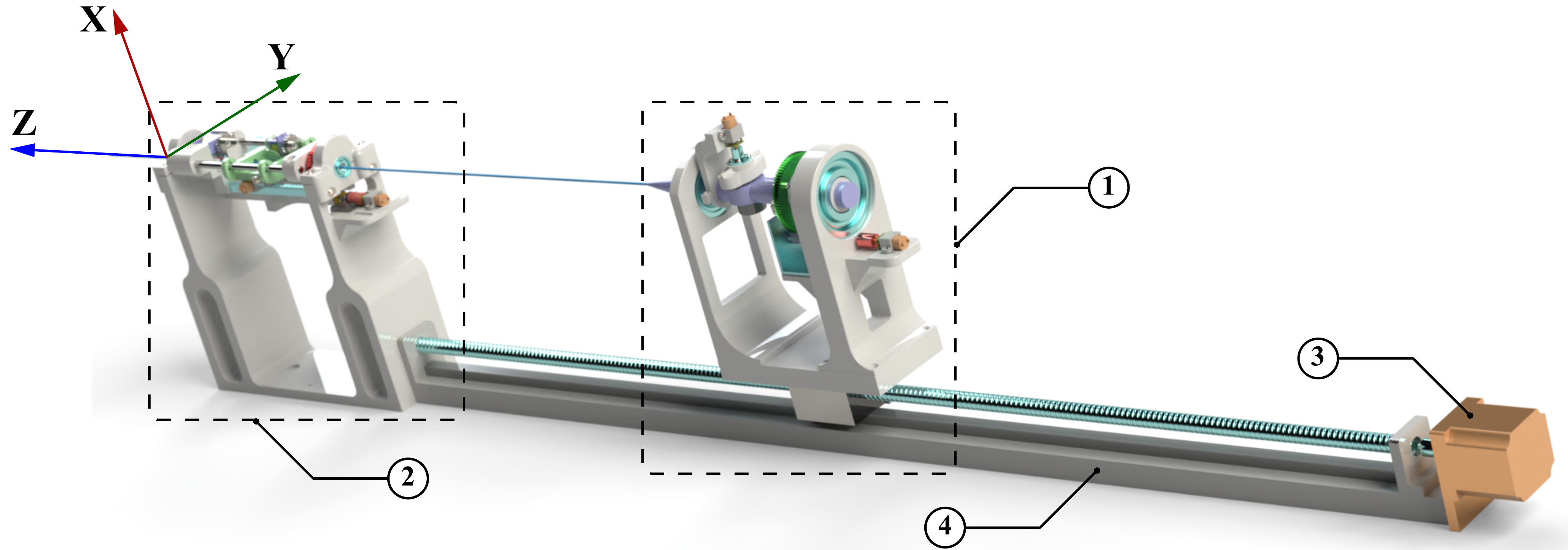}
        \caption{3D CAD model of the follower device: (1) handler mechanism; (2) feeder mechanism; (3) translational motor - handler; (4) linear rail - handler.}
        \label{fig:followerCAD}
    \end{center}
\end{figure*}

\begin{figure*}%[thpb]
    \begin{center}
        \includegraphics[width=0.9\textwidth]{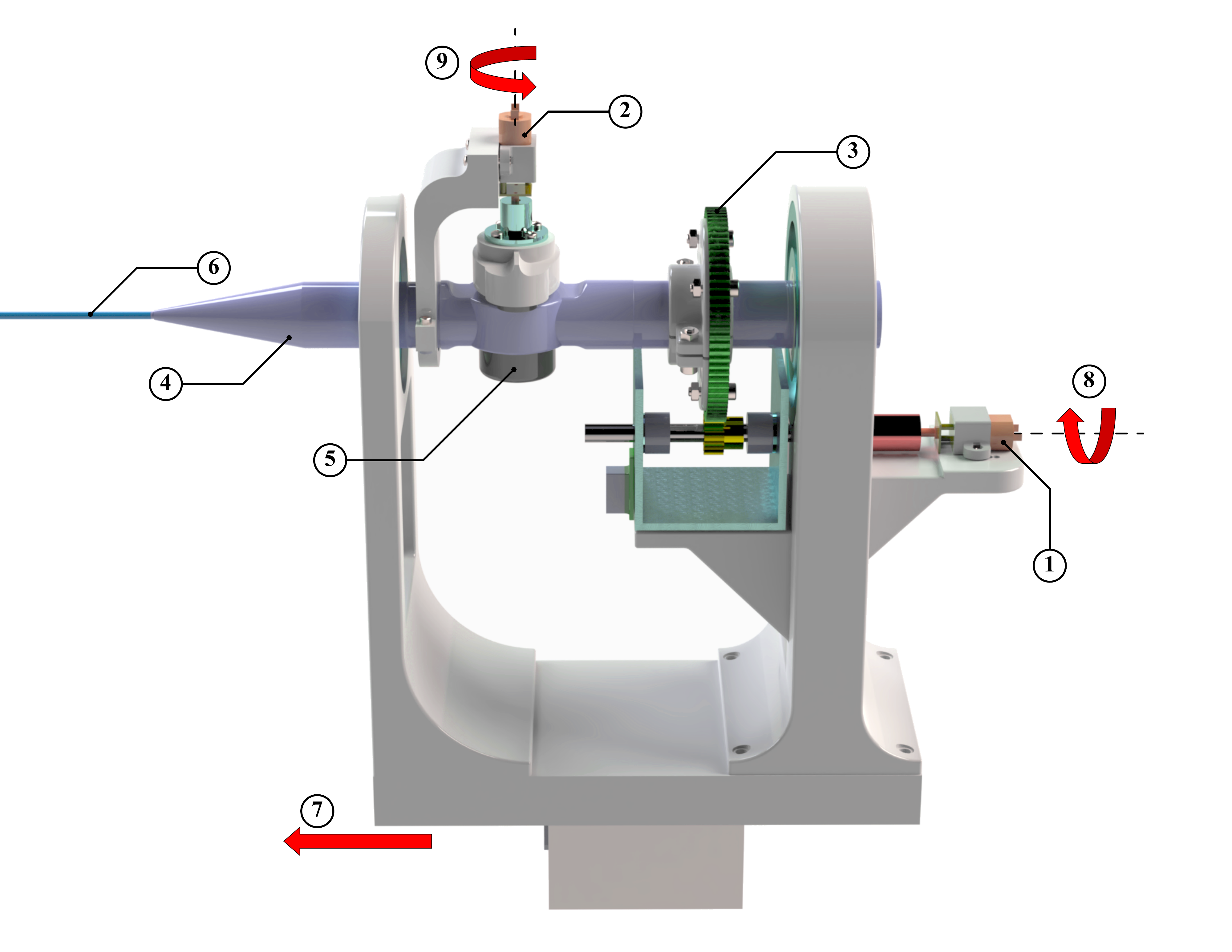}
        \caption{3D CAD model of the follower device's handler: (1) rotational motor; (2) bending motor; (3) rotational gear chain; (4) catheter handle; (5) bending knob; (6) catheter shaft; (7) direction of translation; (8) axis of rotation - rotational motor; (9) axis of rotation - bending motor.}
        \label{fig:handler}
        
        \includegraphics[width=0.9\textwidth]{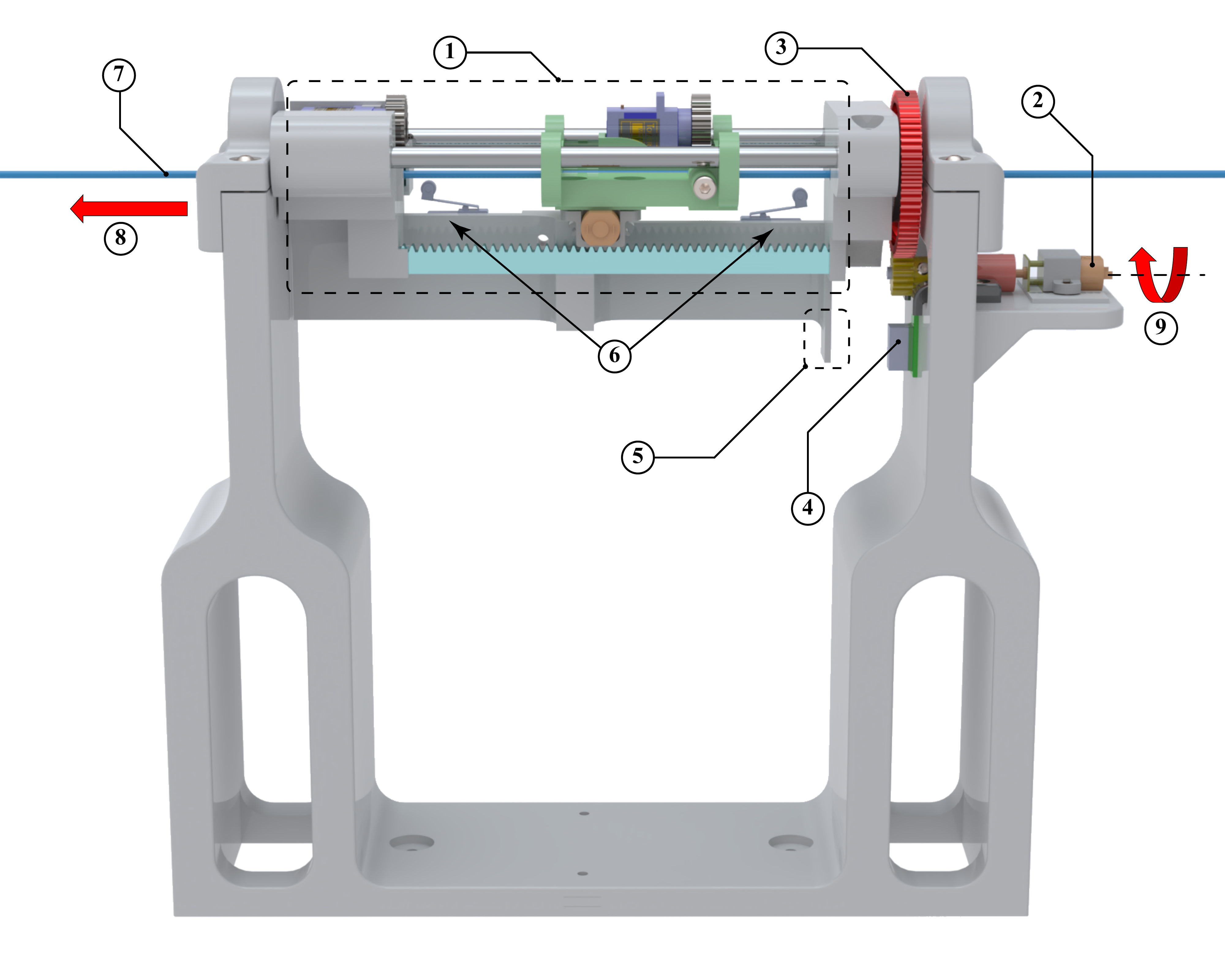}
        \caption{3D CAD model of the follower device's feeder: (1) grip-insert-release assembly; (2) rotational motor; (3) rotational gear chain; (4) proximity sensor; (5) proximity sensor target; (6) limit switches; (7) catheter shaft; (8) direction of translation.}
        \label{fig:feeder}
    \end{center}
\end{figure*}

The follower device, illustrated in Fig. \ref{fig:followerCAD}, consists of handler and feeder mechanisms designed to accommodate a bi-directional navigation ablation catheter. The handler mechanism, shown in Fig. \ref{fig:handler}, holds the catheter handle, while the feeder mechanism, shown in Fig. \ref{fig:feeder}, supports the catheter shaft. Both mechanisms are designed to occupy minimal space in the operating room and ensure precise control. The low-level control architecture for both mechanisms employs a proportional-integral-derivative (PID) controller. The PID controller is implemented by a microcontroller attached to the device. It provides a safe velocity range in all three movements. Similarly to the master device, the strength and internal resistance of the DC motors of the follower device should adapt to the user requirements for device operation. A key distinction is that catheter bending requires a significant amount of force due to direct interaction with the catheter knob. Consequently, a higher gear ratio is necessary, compared to that of the master device, to bend the catheter smoothly and powerfully.

\noindent\textbf{Handler Mechanism:} The handler mechanism comprises three motors responsible for translation, rotation, and bending (see Fig. \ref{fig:handler}). A stepper motor drives translational movement through a ball screw, moving a carriage that holds the catheter handle along with the rotation and bending motors. Rotational movement is facilitated by a DC motor attached to one end of the main handler assembly. This DC motor is mounted on a small platform with a motor bracket, and its shaft, aided by a gearing chain, rotates the catheter handle. The bending of the catheter tip is actuated by another DC motor that turns a knob located on the catheter handle. These three motors are connected to a microcontroller, which provides the necessary signals and controls the movement of the catheter.

\begin{figure*}%[thpb]
    \begin{center}
        \includegraphics[width=.8\textwidth]{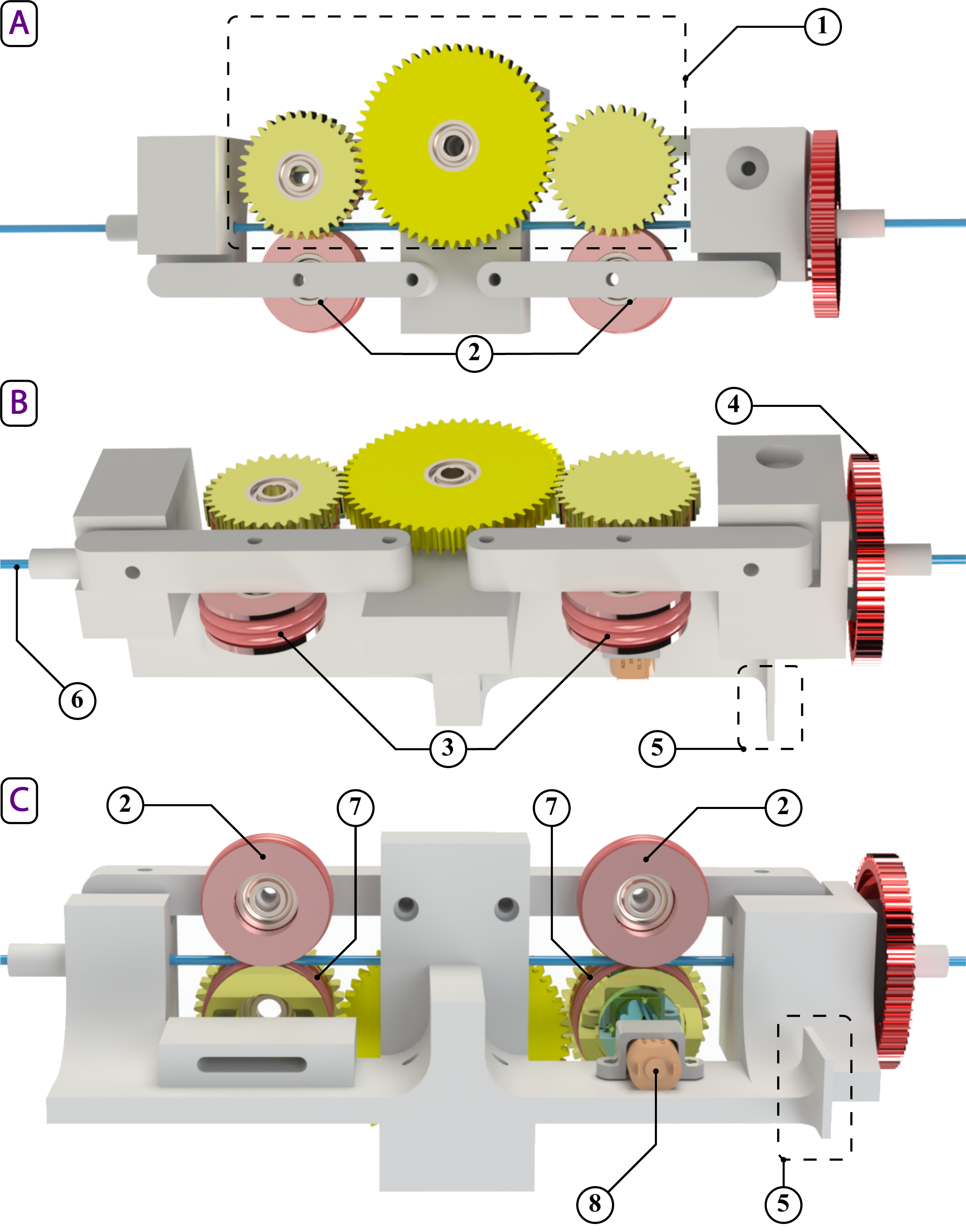}
        \caption{3D CAD model of the initial feeder mechanism featuring a roller-type design from top (A), side (B), and bottom (C) views: (1) linear feeder drive gear chain; (2) gripping idlers; (3) gripping idler O-rings; (4) rotational gear; (5) proximity sensor target; (6) catheter shaft; (7) drive gear O-rings; (8) linear drive gear motor.}
        \label{fig:ORING}
    \end{center}
\end{figure*}

\noindent\textbf{Feeder Mechanism:} The initial feeder mechanism employed a roller-type design (see Fig. \ref{fig:ORING}), which was prevalent in the literature \cite{baek2022design}. This design, however, exhibited unexpectedly high torsion after implementation. Subsequently, a new feeder was designed to mimic the manual techniques used by surgeons, which involved gripping, inserting, and releasing the catheter. To the best of our knowledge, no robotic catheterization system has ever used such a design.
\begin{figure*}%[thpb]
    \begin{center}
        \includegraphics[width=\textwidth]{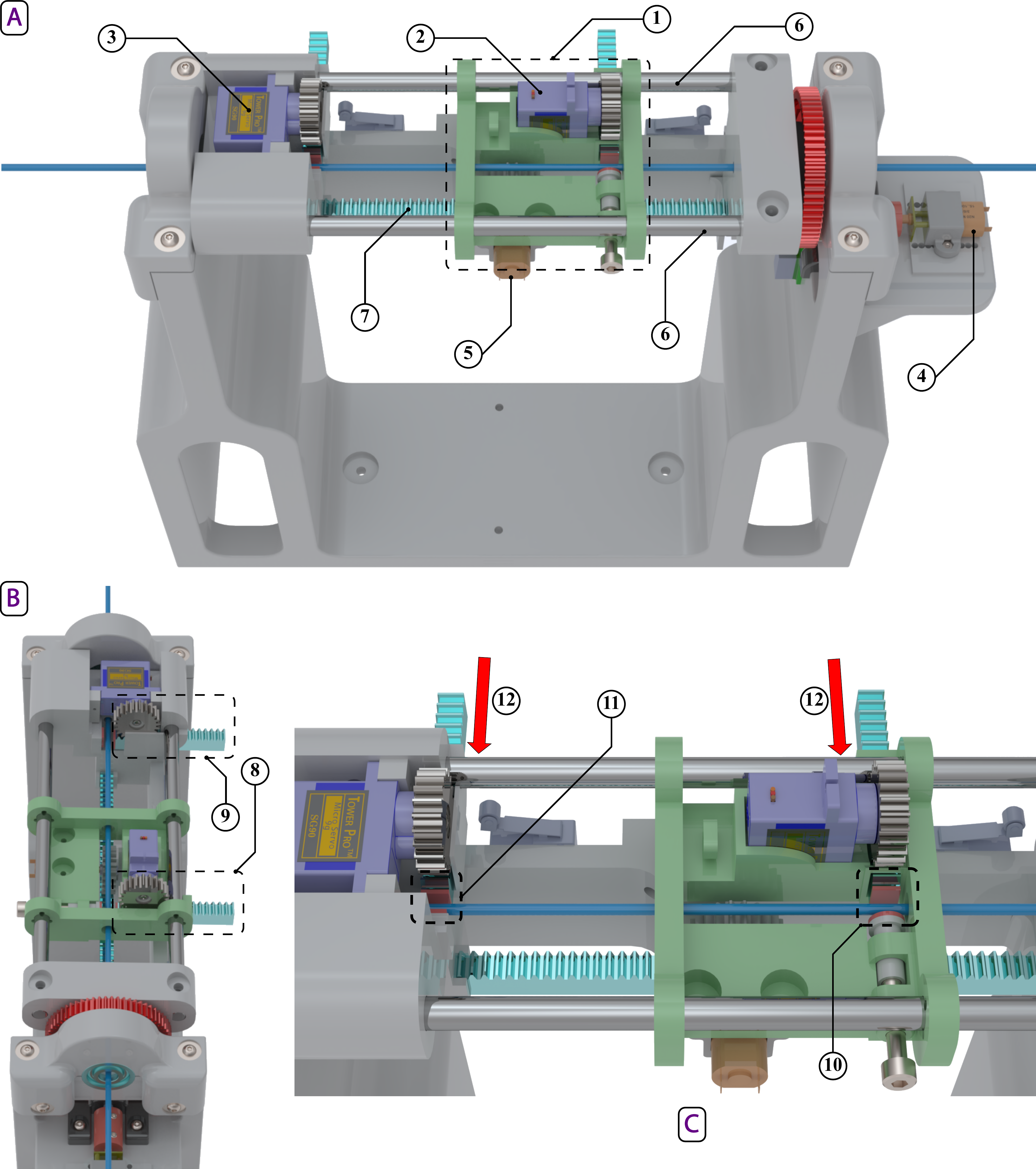}
        \caption{3D CAD model of the final feeder mechanism featuring a grip-insert-release design from side (A), top (B), and zoomed-in (C) views: (1) gripper cart assembly; (2) gripper cart's gripping motor; (3) forward gripping motor; (4) rotational motor; (5) translational motor; (6) gripper cart's rail; (7) translational rack gear; (8) gripping rack gear - gripper cart; (9) gripping rack gear - forward gripper; (10) gripper pads - gripper cart; (11) gripper pads - forward gripper; (12) motion direction - gripping rack gears.}
        \label{fig:GNR}
    \end{center}
\end{figure*}

As shown in Fig. \ref{fig:GNR}, the redesigned feeder mechanism consists of two gripping mechanisms and two DC motors: one for translation and the other for rotation. The gripping mechanisms, used alternately, securely grasp and hold the catheter shaft. The translational component of the feeder mechanism comprises a cart running on two rails. A DC motor, mounted on the underside of the cart, is equipped with a gear that meshes with a stationary rack gear. When actuated, the DC motor drives the cart along the rails, moving parallel to the rack gear. 

The gripping systems are designed to hold the catheter shaft without applying torsion. A servo motor, fitted with a gear, meshes with a shorter rectangular rack gear that moves perpendicular to the catheter shaft. As the rack gear moves, its rubberized gripper pad end applies varying pressure to the catheter shaft. When the rack gear contacts the catheter shaft, it holds the shaft between itself and a stationary rubberized gripper pad mounted on the opposite side of the shaft. One gripping system is located on the cart, while the other is positioned near the catheter tip on a stationary part towards the forward extreme of the feeder. These systems alternate in holding the catheter shaft, with a slight overlap, to minimize buckling. The cart gripper engages when the follower matches the linear movements of the handler, while the forward gripper engages at all other times. 

The translational motor of the feeder must be actuated in synchrony with the stepper motor of the handler mechanism to ensure smooth and buckle-free translation of the catheter tip. Similarly, the rotational DC motor in the feeder mechanism must work synchronously with the handler's rotational movement to minimize torsion on the catheter shaft. The gripper mechanisms also contribute to reducing torsion by applying co-linear forces in the normal direction on both sides of the catheter shaft in a balanced manner. 

%%%%%%%%%%%%%%%%%%%%%%%%%%%%%%%%%%%%%%%%%%%%%%%%%%%%%%%%%%%%%%%
\subsection{Dynamics Modeling} \label{subsec:modeling}

The dynamic model of the proposed system is essential for model-based teleoperation control and for ensuring safe robot-environment interaction. It also aids in catheter characterization, particularly in determining the Young's modulus of the catheter. The mathematical modeling of the follower device includes translation, rotation, and bending. The catheter's bending, under tendon-driven actuation, is modeled using Cosserat rod theory \cite{janabi2021cosserat}. To this end, a customized Cosserat rod-based dynamic model for a single-section continuum robot (i.e., a tendon-driven catheter) is presented. Unlike the generalized model presented in \cite{janabi2021cosserat}, this model is specifically tailored to the 2-DOF tendon-driven catheter used in this study.
\begin{table} %[!thpb]
    \centering
    \caption{Nomenclature for dynamics modeling.}
    \begin{tabular}{c l}
        \toprule  
        \textbf{Symbol} & \textbf{Definition} \\
        \midrule
        $B_{bt}$ & Viscous damping matrix for bending and torsion \\
        $B_{se}$ & Viscous damping matrix for shear and extension \\
        $c_0, c_1, c_2, d_1$ & Coefficients of $BDF-\alpha$ numerical solution method \\
        $K_{bt}$ & Stiffness matrix for bending and rotation \\
        $K_{se}$ & Stiffness matrix for shear and extension \\
        $J$ & Rotational inertia matrix \\
        $\rho$ & Material density \\
        $g$ & Gravity vector \\
        $r^t$ & Tendon's offset from the cross section's center of mass \\
        \botrule
    \end{tabular}
    \label{tab:nomenclature}
\end{table}

Consider that the intended catheter, functioning as a medical continuum robot, consists of an elastic backbone with two continuous channels for routing knob-operated tendons. The rotational displacement of the knob applies tensions (${\tau _i},\,\,i = 1,\,2$) to the tendons, resulting in distributed forces and moments along the length of the backbone. The catheter, viewed as a one-directional rod, is characterized by the centerline curve $p(s,t) \in {\mathbb{R}^3}$ and the material orientation $R(s,t) \in {\mathbb{R}^{3 \times 3}}$, both of which are functions of time $t \in \mathbb{R}$ and an arc length parameter $s \in \mathbb{R}$ along the undeformed rod centerline. 

Considering $v(s,t) = {R^T}\frac{{\partial p}}{{\partial s}}$ as the rate of change of position with respect to arc length, $u(s,t) = {({R^T}\frac{{\partial R}}{{\partial s}})^ \vee } \in {\mathbb{R}^3}$ as the curvature, $q = {R^T}\frac{{\partial p}}{{\partial t}}$ as the linear velocity, and $\omega  = {({R^T}\frac{{\partial R}}{{\partial t}})^ \vee } \in {\mathbb{R}^3}$ as the angular velocity, all in the local frame, the dynamics of the catheter can be represented by
\begin{equation}
  \frac{{\partial p}}{{\partial s}} = Rv
\label{eq1}
\end{equation}
\begin{equation}
  \frac{{\partial R}}{{\partial s}} = R \hat u 
\label{eq2}
\end{equation}
\begin{equation}
  \frac{{\partial q}}{{\partial s}} = \frac{{\partial v}}{{\partial t}} - q + v
\label{eq3}
\end{equation}
\begin{equation}
  \frac{{\partial \omega }}{{\partial s}} = \frac{{\partial u}}{{\partial t}} - \hat u\omega
\label{eq4}
\end{equation}
\begin{equation}
  \left[ {\begin{array}{*{20}{l}}
  {\frac{{\partial v}}{{\partial s}}} \\ 
  {\frac{{\partial u}}{{\partial s}}} 
\end{array}} \right] = {{\left[ {\begin{array}{*{20}{c}}
  {{K_{se}} + {c_0}{B_{se}} + A}&G \\ 
  {{G^T}}&{{K_{bt}} + {c_0}{B_{bt}} + H} 
\end{array}} \right]}^{ - 1}}\left[ {\begin{array}{*{20}{l}}
  {{\Pi _n} - {\Sigma _n}} \\ 
  {{\Pi _m} - {\Sigma _m}} 
\end{array}} \right]
\label{eq5}
\end{equation}
\begin{equation}
  \frac{{\partial p}}{{\partial t}} = Rq
\label{eq6}
\end{equation}
\begin{equation}
  \frac{{\partial R}}{{\partial t}} = R \hat \omega
\label{eq7}
\end{equation}
where 
\begin{equation}
  {\Pi _n} = \rho A(\hat \omega q + \frac{{\partial q}}{{\partial t}}) + C\left[ {\begin{array}{*{20}{c}}
  {q_1^2\operatorname{sgn} ({q_1})}&{q_2^2\operatorname{sgn} ({q_2})}&{q_3^2\operatorname{sgn} ({q_3})} 
\end{array}} \right] - {R^T}\rho Ag - a
\label{eq8}
\end{equation}
\begin{equation}
  {\Pi _m} = \rho (\hat \omega J\omega  + J\frac{{\partial \omega }}{{\partial t}}) - \hat v ({K_{se}}(v - {v^*}) + {B_{se}}\frac{{\partial v}}{{\partial t}}) - b
\label{eq9}
\end{equation}
\begin{equation}
{\Sigma _n} = \hat u  ({K_{se}}(v - {v^*}) + {B_{se}}\frac{{\partial v}}{{\partial t}}) - {K_{se}}\frac{{\partial {v^*}}}{{\partial s}} + {B_{se}}(\frac{{\partial v}}{{\partial s}}({c_1} + {c_2}) + {d_1}\frac{{\partial v}}{{\partial s\partial t}}) 
\label{eq10}
\end{equation}
\begin{equation}
  {\Sigma _m} = \hat u  ({K_{bt}}(u - {u^*}) + {B_{bt}}\frac{{\partial u}}{{\partial t}}) - {K_{bt}}\frac{{\partial {u^*}}}{{\partial s}} + {B_{bt}}(\frac{{\partial u}}{{\partial s}}({c_1} + {c_2}) + {d_1}\frac{{\partial u}}{{\partial s\partial t}}) 
\label{eq11}
\end{equation}
\begin{equation}
  a = \sum\limits_{i = 1}^2 {{a_i}} ,\,\,\,{a_i} = {A_i}\left[ {\hat u (\frac{\partial }{{\partial s}}p_i^b + \frac{\partial }{{\partial s}}r_i^t) + \frac{{{\partial ^2}}}{{\partial {s^2}}}r_i^t} \right] 
\label{eq12}
\end{equation}
\begin{equation}
  b = \sum\limits_{i = 1}^2 {{b_i}} ,\,\,\,{b_i} = \hat {r_i^t} {a_i} 
\label{eq13}
\end{equation}
\begin{equation}
  A = \sum\limits_{i = 1}^2 {{A_i},\,\,\,{A_i} =  - {\tau _i}\frac{{{{({\hat{(\frac{\partial }{{\partial s}}p_i^b)} })}^2}}}{{{{\left\| {\frac{\partial }{{\partial s}}p_i^b} \right\|}^3}}}} 
\label{eq14}
\end{equation}
\begin{equation}
  G = \sum\limits_{i = 1}^2 {{G_i}} ,\,\,\,{G_i} =  - {A_i} \hat{r_i^t} 
\label{eq15}
\end{equation}
\begin{equation}
  H = \sum\limits_{i = 1}^2 {\hat{r_i^t} {G_i}} 
\label{eq16}
\end{equation}
\begin{equation}
  \frac{\partial }{{\partial s}}p_i^b = \hat u r_i^t + \frac{\partial }{{\partial s}}r_i^t + v 
\label{eq17}
\end{equation}
Note that ${(.)^ \vee }:g \in {\mathbb{R}^{3 \times 3}} \to {\mathbb{R}^3}$ while $\hat {(.)} :{\mathbb{R}^3} \to g \in {\mathbb{R}^{3 \times 3}}$, Here, $v^*$ and $u^*$ are the initial values of $v$ and $u$, respectively, and $(.)^b$ denotes the variable represented in the local frame. For a typical vector $V = [\alpha \, \beta \, \gamma]^T$,
\begin{equation}
  (\hat V) = \left[ {\begin{array}{*{20}{c}}
  0&{ - \gamma }&\beta  \\ 
  \gamma &0&{ - \alpha } \\ 
  { - \beta }&\alpha &0 
\end{array}} \right]; \,\,\, {(\hat V)^ \vee = V.}\ 
\label{eq18}
\end{equation}
The remaining parameters that appear in the above equations are defined in Table \ref{tab:nomenclature}. To transform Eq. \ref{eq1} into a set of ordinary differential equations, a semi-discretization approach in the time domain, called $BDF-\alpha$, is implemented. Where necessary, the forward Euler method is used for spatial integration \cite{janabi2021cosserat}. 

After deriving the model of the continuum part of the follower system, the translational and rotational movements of the system should be modeled. Since the linear and angular motions of the follower system in target tasks are reasonably slow, the translation and rotation DOFs are modeled quasi-statically. Consequently, the complete motion of the follower system can be described.

%%%%%%%%%%%%%%%%%%%%%%%%%%%%%%%%%%%%%%%%%%%%%%%%%%%%%%%%%%%%%%%
\subsection{Catheter Characterization} \label{subsec:characterization}

In order to control the interaction of the catheter tip with the environment in realistic robotic catheterization, the dynamics model derived in Section \ref{subsec:modeling} is utilized. Additionally, understanding Young's modulus of the catheter, along with its geometrical characteristics and material properties such as shaft length in the active section, external diameter of the shaft, and catheter density, is crucial. Young's modulus can be determined through static or dynamic methods. Dynamic identification involves attaching an external load to the catheter tip and observing the oscillations around a reference point after the load is removed. The peaks of these oscillations are used to extract Young's modulus under the assumption of fully elastic behavior. 

The THERMOCOOL SMARTTOUCH\textsuperscript{\textregistered} SF catheter (Biosense Webster Inc, CA, USA) was used in this study, exhibiting partial relaxation rather than perfect elasticity, thus necessitating static identification methods. Geometric characteristics of the catheter are detailed in Table \ref{tab:catheter_geo}. These numerical values were experimentally determined in the lab by disassembling a catheter identical to the one used in this study.
\begin{table}%[b]
    \centering
    \caption{Geometric characteristics or material properties of catheter.}
    \begin{tabular}{l c l}
        \toprule  
        \textbf{Characteristic} & \textbf{Symbol} & \textbf{Value} \\
        \midrule
        Active length of catheter & $L$ & $0.08\, (m)$ \\
        Outer diameter & $d_o$ & $0.002667\, (m)$ \\
        Second moment of area & $I$ & $1.9165E-12\, (m^4)$ \\
        Density & $\rho$ & $1630.573\, (Kg/m^3)$ \\
        Stiffness & $K$ & $3.01\, (N/m)$ \\
        \botrule
    \end{tabular}
    \label{tab:catheter_geo}
\end{table}

For static identification, only the active section of the catheter shaft was considered, anchored to the feeder mechanism using hot glue to ensure that tip movement was exclusively influenced by external loading. Ten different weights, including an S-shaped hook affixed to the catheter tip with cotton thread, were chosen for loading. Before and after each loading, the initial and displaced positions of the catheter tip were recorded using a C920x (Logitech, Switzerland) HD Pro webcam (1080p/30fps) positioned perpendicular to the catheter's movement plane. The unloading process was conducted carefully, with the loaded hook unloaded step-by-step from the heaviest to the lightest weight, capturing hysteresis effects. Throughout each recording, efforts were made to maintain the stability of the load, ensuring no bouncing or twisting movements occurred. 

Captured frames were processed using OpenCV in Python to convert tip positions from image space to Cartesian coordinates of the follower device. Fig. \ref{fig:lapse} illustrates a time-lapse of the loading-unloading process: frame (0) shows the initial state of the catheter before loading, frames (1)-(9) depict the loading process, and frame (10) represents the maximum loading, which is also the initial state for unloading. Frames (11)-(20) show the unloading sequence, corresponding to frames (9)-(0). Frames (0) and (20) represent the unloaded states before and after the loading-unloading process, respectively. Table \ref{tab:characterization} shows the results of the loading-unloading experiments.
\begin{figure*}%[!t]
    \centering
    \includegraphics[width=\textwidth]{Fig8.pdf}
    \caption{Time-lapse of loading-unloading in catheter characterization: Frame (0) shows the initial state of the catheter; Frames (1)-(9) show the loading; Frame (10) shows the maximum loading, also the initial state for unloading Frames (11)-(20) show unloading, corresponding to frames (9)-(0); Frames (0) and (20) are unloaded cases, respectively, before and after loading.}
    \label{fig:lapse}
\end{figure*}

\begin{table*}%[!t]
    \footnotesize
    \centering
    \captionsetup{width=\textwidth}
    \caption{Catheter characterization results.}
    \begin{minipage}{\textwidth}
    \centering
    \begin{tabular}{>{\centering\arraybackslash}m{1cm} | >{\centering\arraybackslash}m{2cm} >{\centering\arraybackslash}m{2cm} >{\centering\arraybackslash}m{3cm} >{\centering\arraybackslash}m{3cm}}
        \toprule \toprule  
        \textbf{Test case} & \textbf{Weight (gr)} & \textbf{External force (N)} & \textbf{Tip position in loading (mm)} & \textbf{Tip position in unloading (mm)} \\
        \midrule
        0 & 0 & 0 & 0 & 7.79 \\
        \midrule
        1 & 5.08 & 0.0498 & 25.32 & 33.11 \\
        \midrule
        2 & 6.47 & 0.0635 & 31.65 & 37.98 \\
        \midrule
        3 & 7.91 & 0.0776 & 37.01 & 41.88 \\
        \midrule
        4 & 9.32 & 0.0914 & 41.39 & 45.77 \\
        \midrule
        5 & 10.75 & 0.1056 & 44.80 & 48.21 \\
        \midrule
        6 & 12.18 & 0.1195 & 47.72 & 50.64 \\
        \midrule
        7 & 13.59 & 0.1333 & 51.13 & 53.08 \\
        \midrule
        8 & 15.02 & 0.1473 & 53.08 & 55.03 \\
        \midrule
        9 & 16.45 & 0.1614 & 56.00 & 56.49 \\
        \midrule
        10 & 17.95 & 0.1761 & 57.95 & 57.95 \\
        \botrule
    \end{tabular}
    \end{minipage}
    \vspace{1em}
    \begin{minipage}{\textwidth}
    \centering
    \begin{tabular}{>{\centering\arraybackslash}m{1cm} | >{\centering\arraybackslash}m{2cm} >{\centering\arraybackslash}m{3cm} >{\centering\arraybackslash}m{3cm} >{\centering\arraybackslash}m{2cm}}
        \toprule \toprule  
        \textbf{Test case} & \textbf{Point-based Young's modulus (MPa)} & \textbf{Point-based simulated tip position in loading (mm)} & \textbf{Trend-based simulated tip position in loading (mm)} & \textbf{Error (\% of active length)} \\
        \midrule
        0 & 0 & 0 & 0 & 0 \\
        \midrule
        1 & 1.968 & 23.27 & 15.88 & 2.56 \\
        \midrule
        2 & 2.005 & 27.78 & 19.76 & 4.84 \\
        \midrule
        3 & 2.097 & 31.24 & 23.48 & 7.21 \\
        \midrule
        4 & 2.209 & 33.90 & 26.83 & 9.36 \\
        \midrule
        5 & 2.354 & 35.59 & 29.90 & 11.51 \\
        \midrule
        6 & 2.504 & 37.09 & 32.75 & 13.29 \\
        \midrule
        7 & 2.607 & 38.70 & 35.26 & 15.54 \\
        \midrule
        8 & 2.776 & 39.95 & 37.58 & 16.41 \\
        \midrule
        9 & 2.882 & 40.72 & 39.67 & 19.10 \\
        \midrule
        10 & 3.038 & 41.57 & 41.66 & 20.47 \\
        \botrule
    \end{tabular}
    \end{minipage}
    \label{tab:characterization}
\end{table*}

\begin{figure*}%[ht]
    \centering
    \includegraphics[width=0.75\columnwidth]{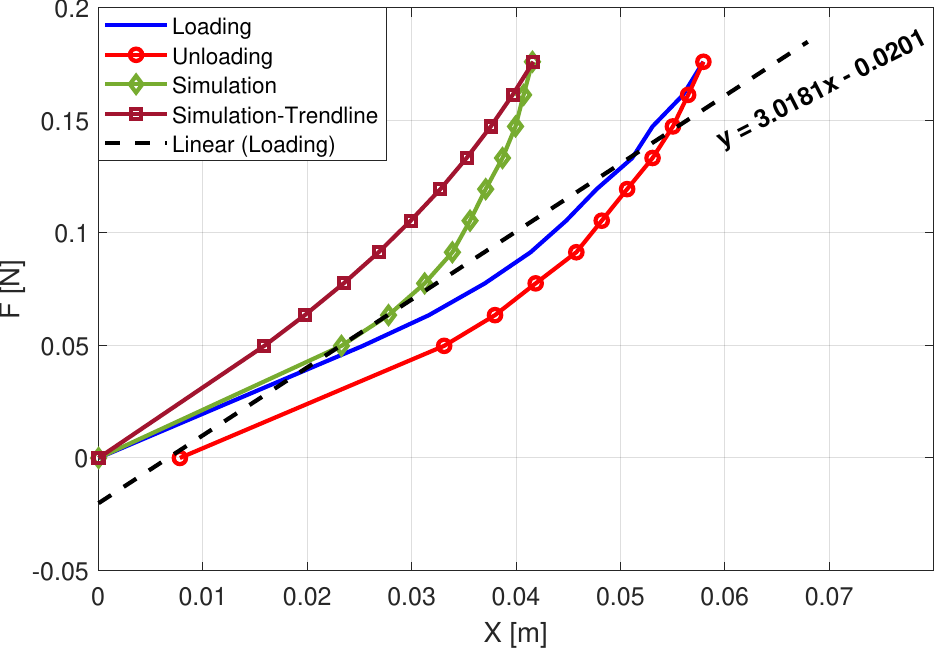}
    \caption{Catheter characterization using force-displacement plots in loading-unloading tests and Cosserat rod-based simulation.}
    \label{fig:characterization}
\end{figure*}

The second column of the table's second half presents point-based Young's modulus calculated by dividing the external force by tip displacement during loading. Additionally, as shown in Fig. \ref{fig:characterization}, Young's modulus was determined by fitting a linear curve to the loading graph, resulting in a calculated modulus of 3.0181 MPa from the slope of the trend line. Both point-based and trend line-based values of Young's modulus were incorporated into the Cosserat rod-based dynamics model of the system (derived in section \ref{subsec:modeling}) to simulate catheter tip positions under specified loading conditions. The third and fourth columns of the table's second half, along with Fig. \ref{fig:characterization}, show simulation results, while the last column shows the percentage error between simulated and experimental loading results relative to the catheter's active length (i.e., 8 cm).

The observed increase in error with increased loading and hysteresis highlights nonlinear behavior in the catheter. Young's modulus of the catheter can be approximated using two linear curves fitted to the first and second halves of the test cases. Alternatively, nonlinear functions such as quadratic curves may be employed to define Young's modulus as a function of external force. These nonlinearities underscore the necessity for a finely tuned closed-loop controller in the follower system to ensure smooth and safe operation.

%%%%%%%%%%%%%%%%%%%%%%%%%%%%%%%%%%%%%%%%%%%%%%%%%%%%%%%%%%%%%%%
\section{Results}

In this section, the following experiments are performed: 1) Approaching five distinct points in the 3D workspace of the follower device by the catheter tip to mimic scenarios common in minimally invasive medical interventions \cite{sikorski2021flexible,rosch2023enhanced}. The entire approaching path is traversed five times to evaluate the repeatability of the designed system (see Section \ref{subsubsec:approaching}); 2) Controlling the follower device in open-loop mode such that the catheter tip tracks circular, infinity-like, and spiral paths in the 3D workspace \cite{tuna2018analysis,jolaei2020design}. Each of the three paths is traversed five times to assess the repeatability of the system (see Section \ref{subsubsec:paths}).

\subsection{Experimental Setup} \label{subsec:setup}

The structural components and the handle assembly of the master device are 3D printed in white polylactic acid (PLA) and, along with the handle track, are mounted onto a base for stability and robustness. PLA parts were printed using an Ender 3 (Creality, China) 3D printer, while the nylon parts were printed using a Form1 (Formlabs, MA, USA) selective laser sintering (SLS) 3D printer. The structural components printed in PLA have 20$\%$ infill, and the SLS nylon parts are full density. These components are designed to be robust, sturdy, and resistant to vibration due to the dynamic response of the motors. To achieve high-resolution movements, the follower device uses a combination of nylon SLS-printed gears and commercially sourced metallic gears, providing a higher number of gear teeth and resolution compared to previous iterations that used PLA-printed gears.

GP2Y0D805Z0F (Sharp, Japan) digital proximity sensors (infrared with a range of 5 cm) were installed on both the handler and feeder mechanisms to assist in the calibration of the rotational frame of reference. All N20 (TT Motor Industrial Co., China) DC motors in the master and follower devices are fitted with 28 pulses per revolution on the encoder side. The encoders assist in calculating the angular position of the motor shaft. Depending on the gear ratios of the gearboxes added to the geared motors, the resolutions on the motor shafts are much higher, as provided in Table \ref{tab:motors}.

For experimentation, the follower device was fitted with a THERMOCOOL SMARTTOUCH\textsuperscript{\textregistered} SF bi-directional navigation catheter (Cat No.: D134804). The catheter has an asymmetric (F-J) tip bending characteristic and an insertion length of 115 mm. The master and follower devices are controlled by Arduino Mega and Due (Arduino AG, Italy) microcontrollers, respectively, both of which are connected to a desktop computer. 

\subsubsection{Master Device} \label{subsubsec:masterPrototype}

\begin{figure*}%[t]
    \begin{center}
        \includegraphics[width=\textwidth]{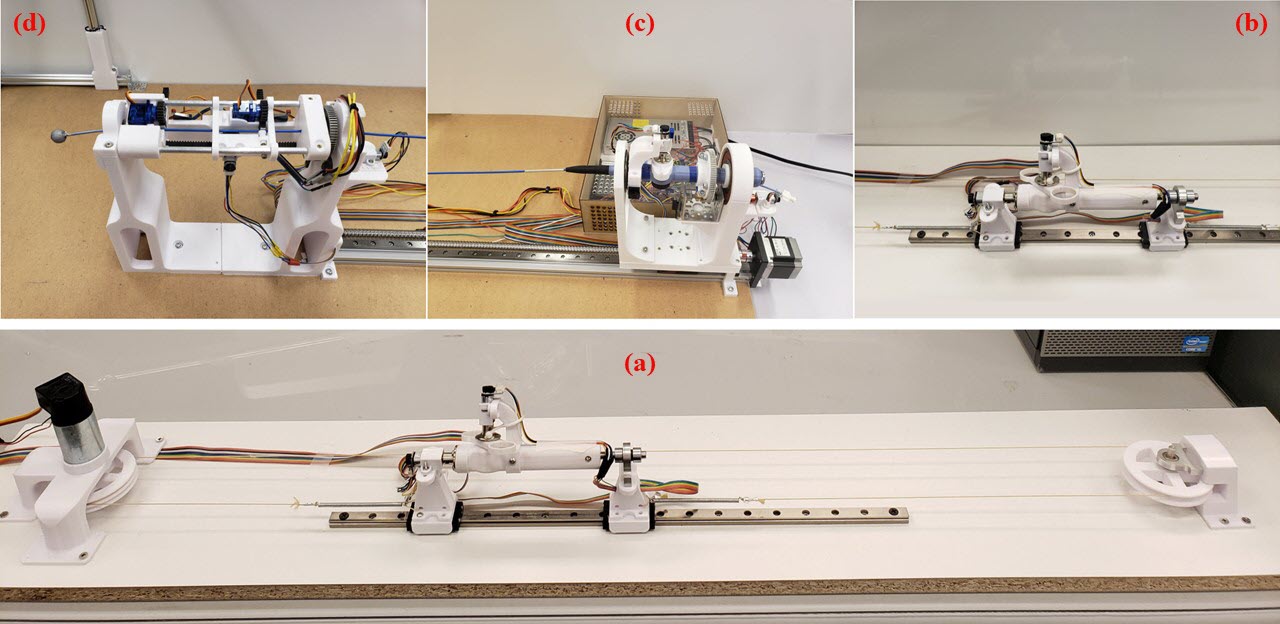}
        \caption{Prototype of master and follower devices: (a) Full view of the master device; (b) Close view of the handle in the master device; (c) Handler mechanism in follower device; (d) Feeder mechanism in follower device, including Vicon marker at the catheter tip.}
        \label{fig:prototype}
    \end{center}
\end{figure*}

The master device is equipped with a foot pedal. The foot pedal engages the translational and rotational systems during experimentation and regular use. The operator can disengage and reengage these systems to facilitate smooth and physically comfortable handle movements within the confines of the linear track. Since the linear range of travel of the master device is approximately one-third that of the follower device, the pedal resets the master without affecting the follower to accommodate longer translations. N20 DC motors were chosen to provide the bending and rotation reaction forces. A gearless Maxon A32 motor (Maxon Motors, Switzerland)  provides the linear reaction force, selected to minimize friction and resistive forces while providing smooth motions. Specific information regarding the motors is detailed in Table \ref{tab:motors}. Fig. \ref{fig:prototype} illustrates the prototyped master device.
\begin{table}%[t!]
    \centering
    \captionsetup{width=\textwidth}
    \caption{Characteristics of DC motors used in the master-follower system.}
    \begin{minipage}{\textwidth}
    \centering
    \begin{tabular}{| >{\arraybackslash}m{2.5cm} | c c c |}
        \toprule \toprule
        \textbf{Device} & \multicolumn{3}{c|}{\textbf{Master}} \\
        \midrule
        \textbf{Motor} & Maxon 32 & N20 & N20 \\
        \midrule
        \textbf{Type} & Brushed & Brushed & Brushed \\
        \midrule
        \textbf{Motion} & Trans. & Rot. & Bend. \\
        \midrule
        \textbf{Int. Gear Ratio} & - & 150:1 & 100:1 \\
        \midrule
        \textbf{Ext. Gear Ratio} & - & - & - \\
        \midrule
        \textbf{Power} [W] & 20 & 0.48 & 0.50 \\
        \midrule
        \textbf{Voltage} [V] & 6 & 6 & 6 \\
        \midrule
        \textbf{No-load Current} [mA] & 123 & 70 & 70 \\
        \midrule
        \textbf{No-load Speed} [RPM] & 4880 & 150 & 220 \\
        \midrule
        \textbf{Stall \,Current} [mA] & 13200 & 670 & 670 \\
        \midrule
        \textbf{Stall Torque} [mNm] & 153 & 127 & 68.6 \\
        \midrule
        \textbf{Resolution on Shaft} [deg.] & 0.72 & 0.086 & 0.13 \\
        \botrule
    \end{tabular}
    \end{minipage}
    \vspace{1em}
    \begin{minipage}{\textwidth}
    \centering
    \begin{tabular}{| >{\arraybackslash}m{2.5cm} | c c c | c c c |}
        \toprule \toprule
        \textbf{Device} & \multicolumn{3}{c|}{\textbf{Follower - Handler}} & \multicolumn{3}{c|}{\textbf{Follower - Feeder}} \\
        \midrule
        \textbf{Motor} & Nema 23 & N20 & N20 & N20 & N20 & 2\,x\,SG90\\
        \midrule
        \textbf{Type} & Stepper & Brushed & Brushed & Brushed & Brushed & Micro Servo \\
        \midrule
        \textbf{Motion} & Trans. & Rot. & Bend. & Trans. & Rot. & Gripper \\
        \midrule
        \textbf{Int. Gear Ratio} & - & 380:1 & 1000:1 & 210:1 & 298:1 & - \\
        \midrule
        \textbf{Ext. Gear Ratio} & - & 21:4 & - & 1.86\footnotemark[1] & 4:1 & 0.804\footnotemark[1] \\
        \midrule
        \textbf{Power} [W] & - & 0.53 & - & 0.46 & 0.44 & - \\
        \midrule
        \textbf{Voltage} [V] & 24 & 6 & 6 & 6 & 6 & 4.8\footnotemark[2] \\
        \midrule
        \textbf{No-load Current} [mA] & - & 70 & 70 & 70 & 70 & - \\
        \midrule
        \textbf{No-load Speed} [RPM] & - & 57 & 22 & 100 & 73 & 100\footnotemark[3] \\
        \midrule
        \textbf{Stall \,Current} [mA] & - & 670 & 670 & 670 & 670 & - \\
        \midrule
        \textbf{Stall Torque} [mNm] & 1000\footnotemark[4] & 353 & 657 & 167 & 245 & 177 \\
        \midrule
        \textbf{Resolution on Shaft} [deg.] & 0.113 & 0.034 & 0.013 & 0.061 & 0.043 & - \\
        \botrule
    \end{tabular}
    \end{minipage}
    \footnotesize \footnotetext{\textsuperscript{1} mm per revolution; \textsuperscript{2} Operating voltage; \textsuperscript{3} Operating speed; \textsuperscript{4} Holding torque}
    \label{tab:motors}
\end{table}

\subsubsection{Follower Device} \label{subsubsec:followerPrototype}

The motors for the handler and the feeder were chosen according to Table \ref{tab:motors}. N20 motors actuate the rotation, bending, and linear motions in the feeder. SG90 (TowerPro, Taiwan) 9g micro servo motors were chosen for the handler gripper motors. Limit switches are attached to the rail at the two extremes of the desired translation of the handler to prevent unwanted linear translations and protect the mechanism from unexpectedly fast movements. The proximity sensors in the handler and feeder synchronize common zero points for the rotational motors and their encoder readings. During experimentation, the rotational position was controlled using the rotational motor encoder readings combined with calibration using the infrared sensors. In the feeder, the pinion gears in the grippers and the translational actuator are M9, 22-tooth, SLS 3D-printed nylon gears. The pinion in the grippers meshes with 50 mm long rack gears, while the pinion in the translational actuator meshes with a 137.8 mm rack gear. Fig. \ref{fig:prototype} illustrates the prototyped follower device. 

\subsubsection{Master-Follower Integration} \label{subsubsec:integration}

In order to integrate the master and follower devices as a teleoperation system, User Datagram Protocol (UDP) was used to transfer commands and status updates via a network. Master and follower devices were connected to a router using LAN cables to ensure feedback from the follower to the master device in bilateral teleoperation would be as close to real-time as possible. The latency for a two-way communication test (sending and then receiving some commands) was measured for 1 million commands at around 1 to 2 ms. The coding outside the Arduino microcontrollers on both the master and follower sides was carried out in Python to create a dynamic, scalable, and adjustable program in any environment and operating system.

\subsubsection{Motions Mapping for Teleoperation} \label{subsubsec:mapping}

The mapping between master and follower devices is fully adjustable in the program to match the operator’s needs. In this work, the mapping was implemented 1-to-1 for translation, rotation, and bending movements, meaning each subtle movement on the master device is replicated on the follower device. However, the bending in the follower device results in a nonlinear orientation of the catheter tip. 
\begin{figure*} %[b]
    \begin{center}
        \includegraphics[width=0.75\columnwidth]{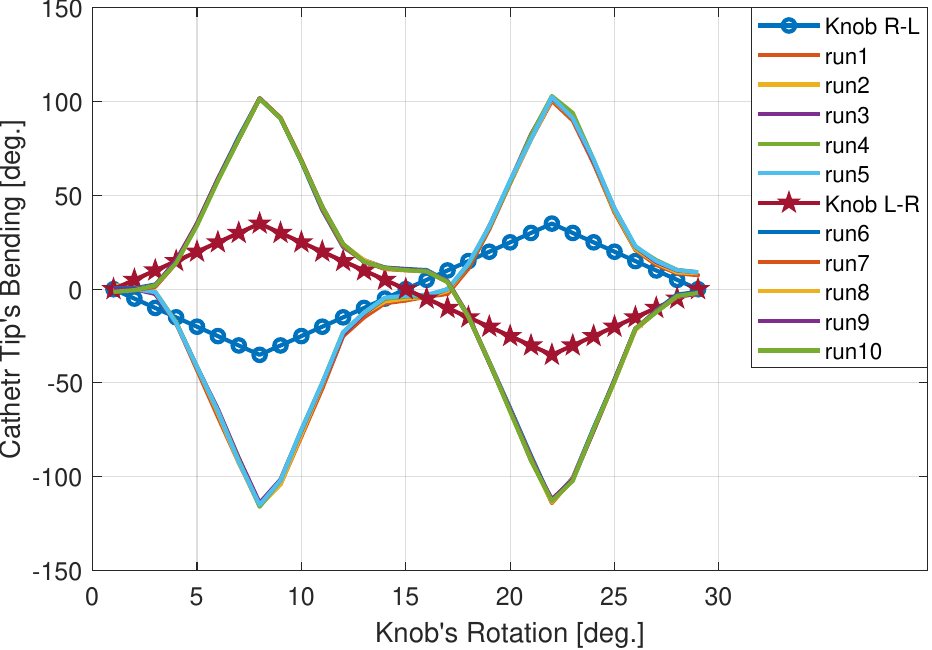}
        \caption{Catheter's bending mapping derived from five left-right-directed (R-L) and five right-left-directed (L-R) bending iterations.}
        \label{fig:bending}
    \end{center}
\end{figure*}

In order to determine the relationship between knob actuation and catheter tip orientation, the knob was actuated incrementally with a resolution of 5 degrees up to 35 degrees, resulting in a catheter tip orientation of at least 90 degrees. Given the asymmetry in the catheter's left-side (F) and right-side (J) bending, experiments were conducted in two modes: (1) right-directed actuation from zero to 35 degrees followed by a return to zero, then left-directed actuation from zero to 35 degrees followed by a return to zero; and (2) left-directed actuation from zero to 35 degrees followed by a return to zero, then right-directed actuation from zero to 35 degrees followed by a return to zero. At each actuation point, the catheter tip orientation was captured using two identical C920x HD Pro webcams (top and side views), and the captured frames were processed to extract the tip orientation. Fig. \ref{fig:bending} presents the obtained results.

As shown in the figure, the catheter exhibits a dead zone between -10 and +10 degrees, where knob actuation does not significantly alter the tip orientation. The catheter's hysteresis is also evident, as it does not return to its initial position after the actuation returns to zero. This hysteresis is observed when the catheter actuation transitions from left-directed to right-directed bending and at the end of the actuation scenario.

\subsection{Open-loop Tracking Control} \label{subsec:control}

Open-loop tracking control is fundamental in medical device teleoperation, particularly in procedures such as catheterization. This control method relies on predefined commands to guide the system without real-time feedback from the catheter's environment. By executing a predetermined path, open-loop control ensures precision and consistency, making it a critical component in the delicate process of navigating catheters through complex vascular pathways. This subsection discusses the implementation of approaching and path tracking over circular, infinity-like, and spiral paths. The approaching scenario mimics what cardiovascular interventionalists practice when navigating a catheter inside the human body to inspect specific points \cite{sikorski2021flexible,rosch2023enhanced}. Circular, infinity-like, and spiral paths are common alternatives in the literature \cite{tuna2018analysis,jolaei2020design} as they can uncover potential dead zones and hysteresis in the system, providing an opportunity for thorough analysis.

\subsubsection{Approaching Scenario} \label{subsubsec:approaching}

In many medical and non-medical tasks, teleoperated robotic systems are used to approach distinct points in 3D space. This task may contain discrete points to be met individually or successively as a series of target points. As a first open-loop control task, the proposed system navigated the catheter tip to five points successively within the workspace of the follower device. In this mode of approaching, a cumulative error can be expected because any error in approaching a point will impact the accuracy of the entire scenario. This task was implemented five times for all target points to explore the repeatability of the system. The catheter tip position was captured using two C920x (Logitech, Switzerland) HD Pro webcams (1080p/30fps), and the tip position was extracted by triangulation. Table \ref{tab:approaching} shows the obtained results. 
\begin{table*}%[t!]
    \centering
    \caption{Approaching results for five distinct points in five cycles.}
    \begin{tabular}{>{\centering\arraybackslash}m{1.5cm} | >{\centering\arraybackslash}m{1.25cm} >{\centering\arraybackslash}m{1.25cm} >{\centering\arraybackslash}m{1.25cm} | >{\centering\arraybackslash}m{1.5cm} >{\centering\arraybackslash}m{1.5cm} >{\centering\arraybackslash}m{1.5cm}}
        \toprule  
        \textbf{Actuation Space Variable (T,R,B)\footnotemark[1]} & \textbf{Mean Value of $X_{w}\, (cm)$} & \textbf{Mean Value of $Y_{w}\, (cm)$} & \textbf{Mean Value of $Z_{w}\, (cm)$} & \textbf{Standard Deviation of $X_{w}$} & \textbf{Standard Deviation of $Y_w$} & \textbf{Standard Deviation of $Z_w$} \\
        \midrule
        (0,0,0) & 24.517 & -3.587 & 38.768 & 1.61 & 0.77 & 0.86 \\
        (55,0,25) & 63.384 & -19.983 & 52.150 & 21.33 & 7.74 & 11.19 \\
        (10,-90,0) & 34.890 & -7.884 & 38.851 & 12.87 & 6.37 & 4.64 \\
        (0,0,-50) & 42.903 & -10.086 & 43.776 & 37.32 & 16.78 & 15.68 \\
        (20,90,0) & 25.702 & -2.331 & 35.125 & 0.96 & 0.57 & 0.27 \\
        (25,90,0) & 31.448 & -5.771 & 35.342 & 6.65 & 3.89 & 1.64 \\
        \botrule
    \end{tabular}
    \footnotesize \footnotetext{\textsuperscript{1} T, R, and B stand for translation (mm), rotation (deg.), and bending (deg.), respectively.}
    \label{tab:approaching}
\end{table*}

\subsubsection{Circular, Infinity-like, and Spiral Path Tracking Scenarios} \label{subsubsec:paths}

\begin{figure*}%[!t]
    \centering
    \begin{subfigure}[t]{0.32\textwidth}
        \centering
        \includegraphics[width=\columnwidth]{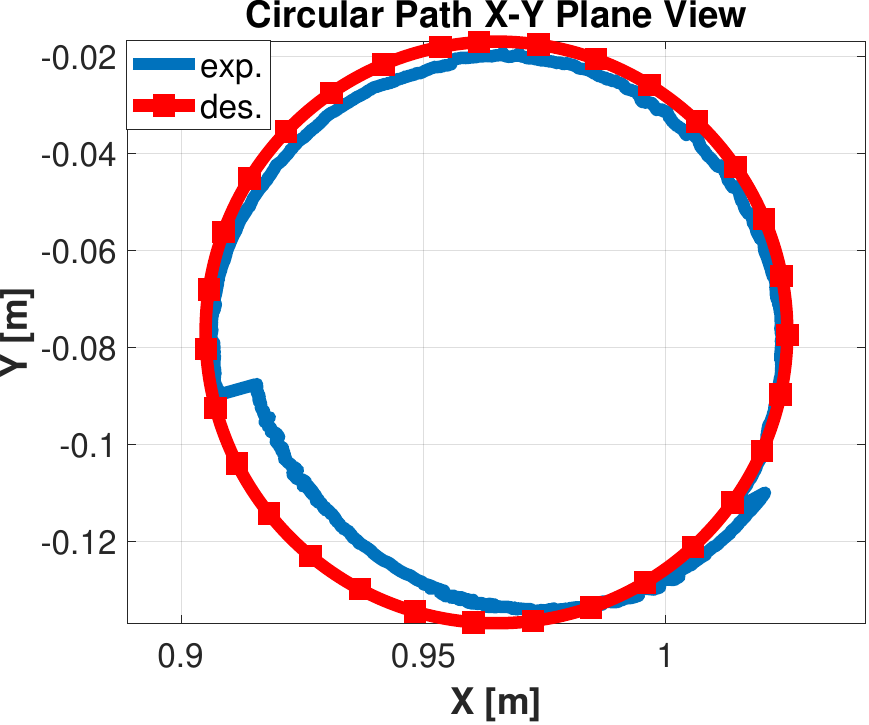}
        \caption{}
    \end{subfigure}%
    ~
    \begin{subfigure}[t]{0.32\textwidth}
        \centering
        \includegraphics[width=\columnwidth]{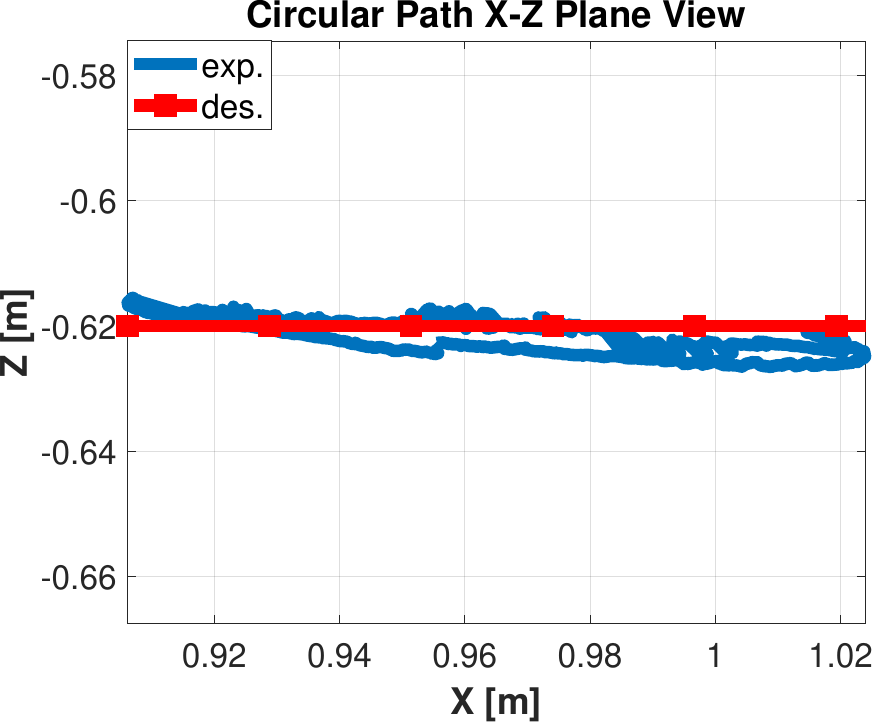}
        \caption{}
    \end{subfigure}%
    ~
    \begin{subfigure}[t]{0.32\textwidth}
        \centering
        \includegraphics[width=\columnwidth]{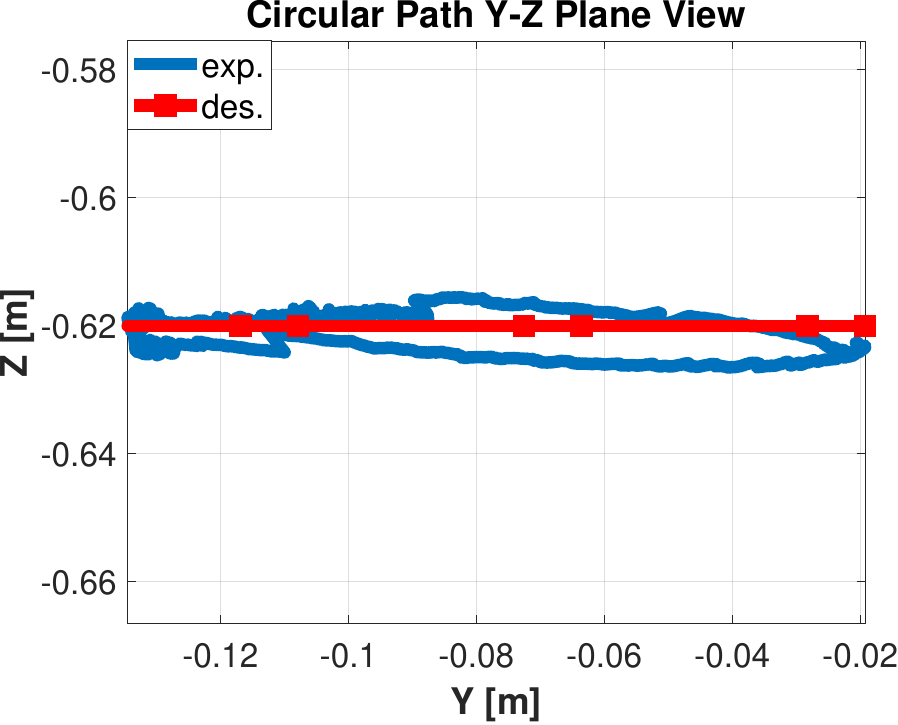}
        \caption{}
    \end{subfigure}%
    \\
    \begin{subfigure}[t]{0.32\textwidth}
        \centering
        \includegraphics[width=\columnwidth]{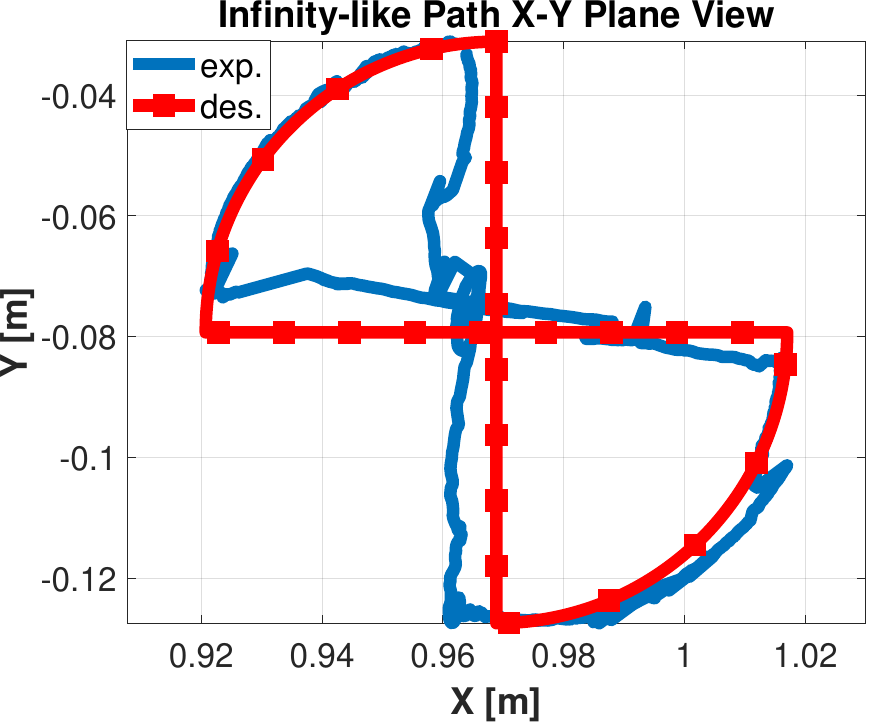}
        \caption{}
    \end{subfigure}%
    ~
    \begin{subfigure}[t]{0.32\textwidth}
        \centering
        \includegraphics[width=\columnwidth]{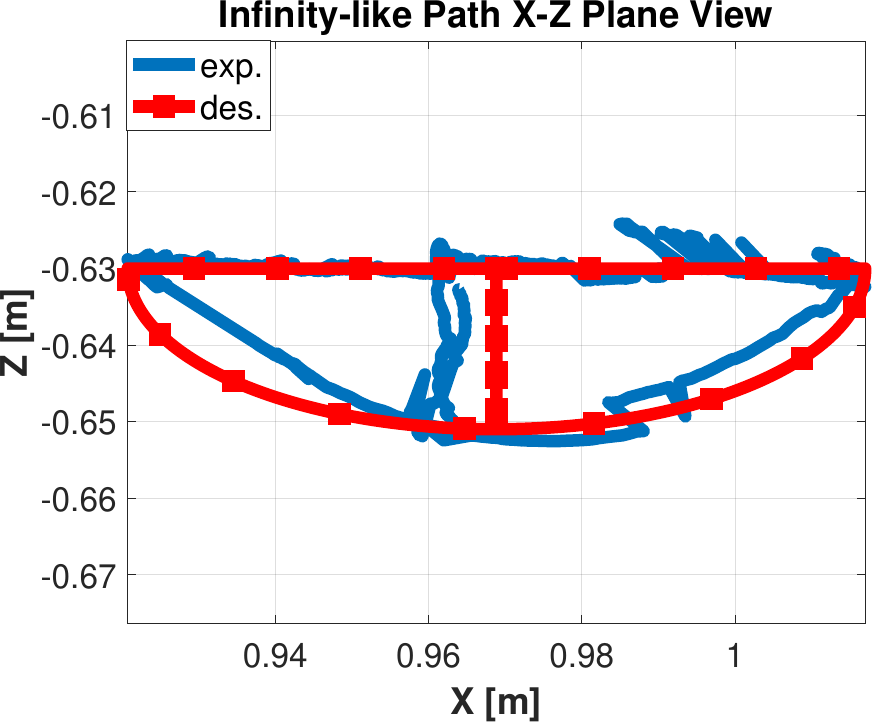}
        \caption{}
    \end{subfigure}%
    ~
    \begin{subfigure}[t]{0.32\textwidth}
        \centering
        \includegraphics[width=\columnwidth]{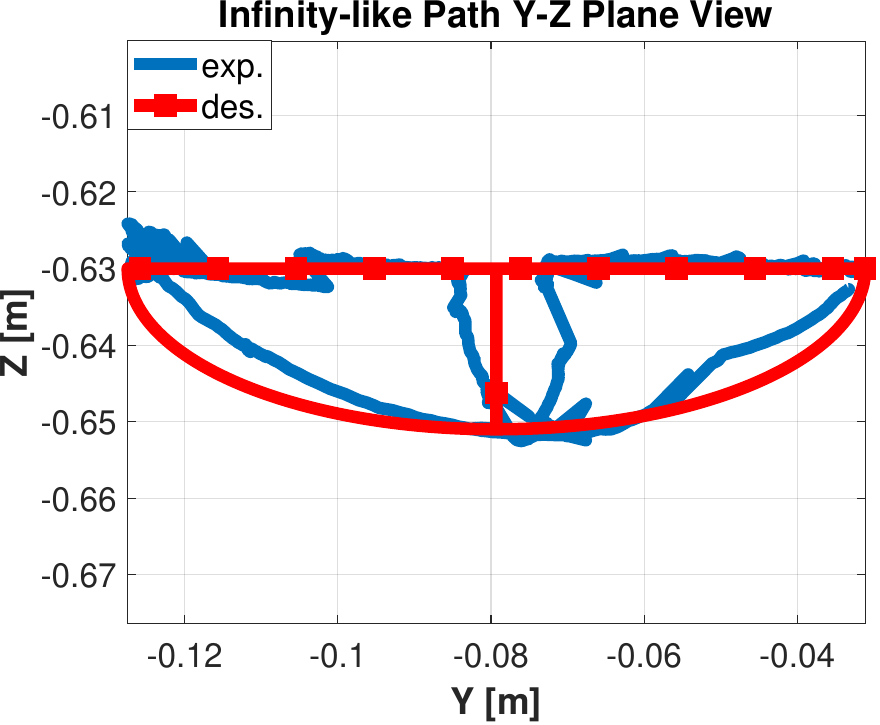}
        \caption{}
    \end{subfigure}%
    \\
    \begin{subfigure}[t]{0.32\textwidth}
        \centering
        \includegraphics[width=\columnwidth]{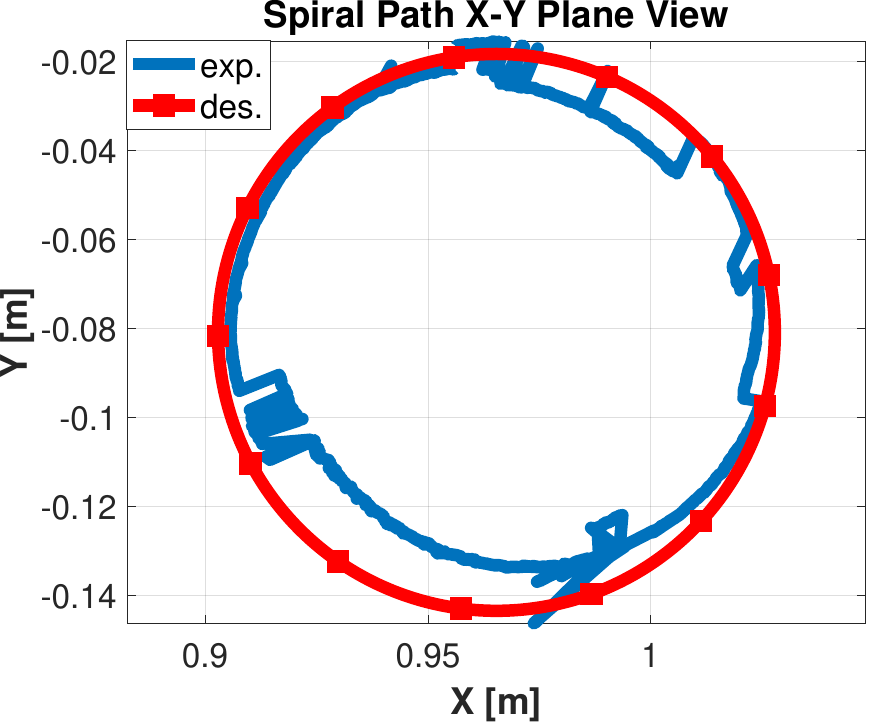}
        \caption{}
    \end{subfigure}%
    ~
    \begin{subfigure}[t]{0.32\textwidth}
        \centering
        \includegraphics[width=\columnwidth]{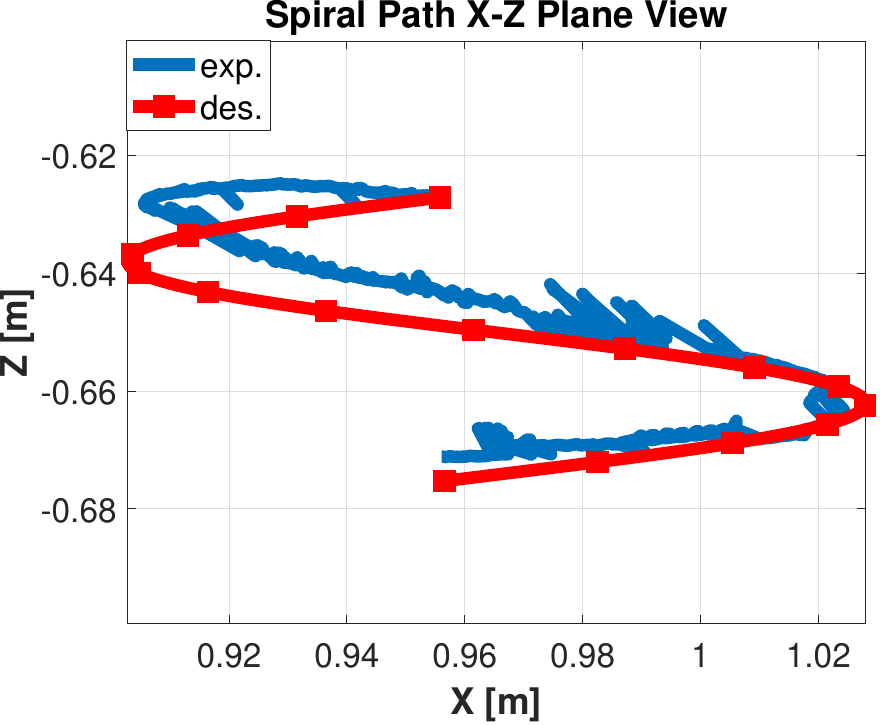}
        \caption{}
    \end{subfigure}%
    ~
    \begin{subfigure}[t]{0.32\textwidth}
        \centering
        \includegraphics[width=\columnwidth]{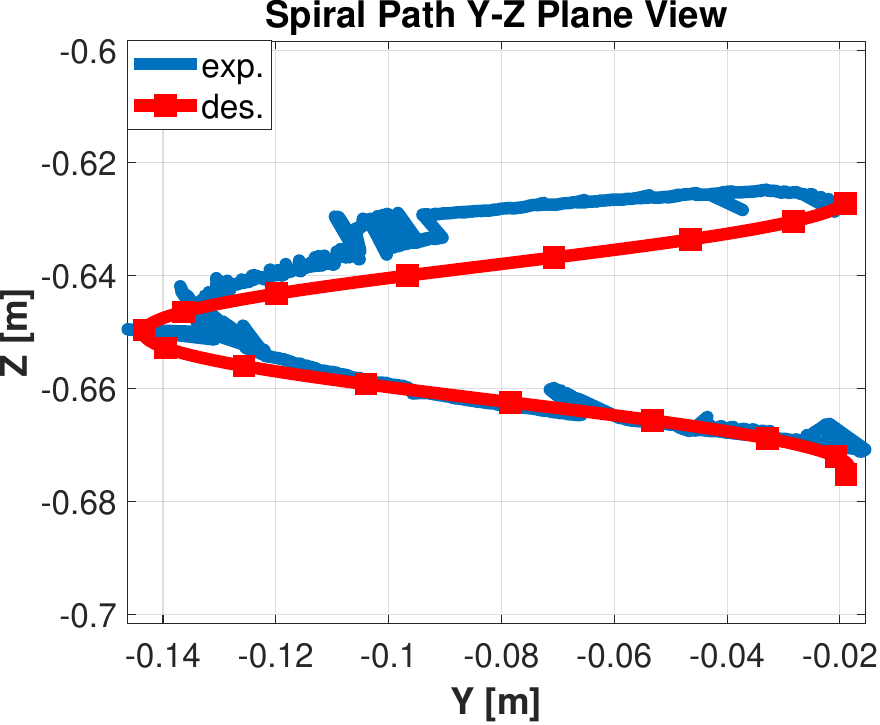}
        \caption{}
    \end{subfigure}%
    \caption{Open-loop path tracking control results in x-y, x-z, and y-z planes: (a)-(c) circular path, (d)-(f) infinity-like path, and (g)-(i) spiral path.}
    \label{fig:OL}
\end{figure*}

In the second phase of the open-loop control tasks, the follower device navigated the catheter tip to track circular, infinity-like, and spiral paths. Each path was tracked five times to assess the system's repeatability. Rather than using webcams to measure catheter tip positions, a Vicon motion capture system (Vicon Motion Systems, UK) was used to increase measurement precision. Motion capture systems typically require a minimum of three markers to determine an object's pose through triangulation. However, attaching three markers to the catheter tip would cause significant bending due to their weight. Consequently, a single marker was utilized, as shown in Fig. \ref{fig:prototype}, to determine the tip's pose. This was achieved using the Object Tracker library \cite{preiss2017crazyswarm}, which processes point cloud data from the motion capture system to provide the pose of each individual marker. The marker used in this experiment weighed 2 gr and had a radius of 16 mm. Pose data was obtained from the motion capture system at a rate of 250 Hz. Fig. \ref{fig:OL} presents sample results from five iterations for three paths.

\section{Discussion} \label{sec:discuss}
\subsection{Design Justification}
Our experimental work with both roller-type and grip-insert-release mechanisms revealed fundamental differences in their mechanical behavior. The roller-type mechanism's continuous contact resulted in complex force distribution along the catheter shaft, where slight unintentional misalignments between rollers, due to manufacturing deficiencies, could compound into significant torsional effects. In contrast, the grip-insert-release mechanism's sequential operation created discrete, well-controlled force applications that proved more predictable and manageable. Analysis of the path tracking results suggests that this difference in force application directly impacted system performance, particularly evident in the infinity-like path results where torsional effects would have most severely impacted performance. The success of this biomimetic approach, based on natural manipulation techniques, suggests potential applications in other medical robotic systems where precise control of flexible instruments is crucial. This new design has not been previously proposed in the literature and represents one of the novel contributions of this work.

\subsection{Tracking Control Results} \label{subsec:controlDisc}

Figure \ref{fig:OL} (a)-(c) illustrates a full cycle of open-loop tracking control for a circular path. A full cycle includes rotation from 0 to 360 degrees followed by a rotation from 360 to 0 degrees to analyze the catheter's hysteresis effect. In this scenario, the catheter was initially translated 55 mm forward, and then the catheter tip was bent 25 degrees. This initial translation ensured that the entire active length of the catheter was free from the feeder mechanism, allowing unrestricted bending. The 25-degree bend angle ensured that the catheter's movements incorporated potential dead zones and hysteresis, facilitating realistic testing. The Vicon system captured the tip position during a full rotation. The imperfections shown in Fig. \ref{fig:OL} (a) are attributed to the wiring of the feeder mechanism, which exerts a loading effect on the feeder's motion, compounded by the gravity effect on the catheter. Figure \ref{fig:OL} (c) demonstrates the gravity effect on the catheter tip, observed in the y-z plane as a mismatch between the desired and experimental paths. The mean Euclidean error (MEE) and mean absolute error (MAE) calculated in the x-y, x-z, and y-z planes for this scenario are listed in Table \ref{tab:errors}.
\begin{table*}%[t!]
    \scriptsize
    \centering
    \caption{Mean Euclidean and mean absolute errors in x-y, x-z, and y-z planes for three path tracking experiments.}
    \begin{tabular}{l c >{\centering\arraybackslash}m{4cm} >{\centering\arraybackslash}m{4cm}}
        \toprule  
        \textbf{Path} & \textbf{Plane} & \textbf{Mean Euclidean Error (cm)} & \textbf{Mean Absolute Error (cm)} \\
        \midrule
        \textbf{Circular} & X-Y & 1.09 & 1.34 \\
         & X-Z & 0.86 & 1.05 \\
         & Y-Z & 0.81 & 1.00 \\
        \midrule
        \textbf{Infinity-like} & X-Y & 0.64 & 0.81 \\
         & X-Z & 0.83 & 0.94 \\
         & Y-Z & 0.89 & 0.99 \\
        \midrule
        \textbf{Spiral} & X-Y & 1.53 & 1.92 \\
         & X-Z & 1.14 & 1.46 \\
         & Y-Z & 1.33 & 1.71 \\
        \botrule
    \end{tabular}
    \label{tab:errors}
\end{table*}

For a full cycle of an infinity-like path, as demonstrated in Fig. \ref{fig:OL} (d)-(f), the catheter was initially translated 55 mm forward, and the catheter tip was bent 30 degrees. The tip was then rotated 90 degrees (subpath 1), bent -60 degrees (subpath 2), rotated -90 degrees (subpath 3), and finally bent 30 degrees (subpath 4). To complete a full cycle, the catheter tip spanned from subpath 4 to subpath 1 successively while the Vicon captured the catheter tip position. The mismatch observed in Fig. \ref{fig:OL} (a) between the real and expected paths in the x-y plane clearly indicates the weighting effect of the catheter and the attached marker. Additionally, Fig. \ref{fig:OL} (c) shows the catheter's dead zone effect (previously observed in Fig. \ref{fig:bending}) in practice, as the vertical segment of the expected path was followed by two different paths in practice, one for the forward (subpaths 1-4) and one for the backward (subpaths 4-1) motion directions. The MEE and MAE calculated in the x-y, x-z, and y-z planes for this scenario are listed in Table \ref{tab:errors}.

Lastly, the master-follower system navigated the catheter tip in a full-cycle spiral path, as shown in Fig. \ref{fig:OL} (g)-(i), by actuating the translation and rotation DOFs. Initially, the catheter was translated 55 mm forward, and the catheter tip was bent 30 degrees. This state of the catheter tip, combined with zero rotation, was considered the initial state. The catheter was then commanded to move 45 mm forward while rotating 360 degrees simultaneously, followed by a backward motion of 45 mm in translation and -360 degrees in rotation to complete a full-cycle motion. Figure \ref{fig:OL} (i) clearly shows both the gravity effect on the catheter and marker in addition to the catheter's hysteresis. The former appears as a rotation of the curve about the axis perpendicular to the x-z plane, while the latter manifests as a two-path plot in practice. The MEE and MAE calculated in the x-y, x-z, and y-z planes for this scenario are listed in Table \ref{tab:errors}.

\subsection{Limitations and Implications of Tracking Control} \label{subsec:LimitImply} 

The results demonstrated in Fig. \ref{fig:OL} reveal factors contributing to errors. 
Dead zone and hysteresis of the catheter are prominent in all experiments. Due to extensive testing, the hysteresis and memory effects became more pronounced. This issue is uncommon in real-world medical interventions since catheters are primarily single-use devices due to biohazards and safety requirements. The gravity effect on the catheter and the marker attached to the catheter tip also contributed to errors. This effect introduces initial errors by making the start points of the desired and experimental paths different. For example, it causes out-of-plane movements in the circular path, where only in-plane movements are expected. The asymmetric geometry of the catheter, uneven friction, and inconsistent movements at different rotational angles and bending also contribute to errors. Systematic errors due to the loading effects of the feeder mechanism wiring, especially in 360-degree rotations, are additional sources of error. The revised tracking algorithm for the single-marker architecture, originally developed for a three-marker setup, introduced some errors despite functioning effectively in our research. Finally, open-loop tracking control does not allow for corrective commands to the motors, making real-time tracking more susceptible to errors. Future research should aim to eliminate one or more of these error sources to achieve smoother and more accurate catheter movements.

%%%%%%%%%%%%%%%%%%%%%%%%%%%%%%%%%%%%%%%%%%%%%%%%%%%%%%%%%%%%%%%
\section{Conclusions} \label{sec:concl}
This paper detailed the design and development of a haptic-enabled teleoperated system for robotic catheterization, incorporating both master and follower devices with translation, rotation, and bending DOFs. The follower device included an innovative grip-insert-release mechanism to prevent catheter buckling and torsion, mimicking manual interventions by clinicians. The system utilized an ablation catheter, which was characterized using a vision-based loading-unloading experiment. Performance evaluations involved approaching and open-loop path tracking along circular, infinity-like, and spiral paths. The results indicated that the system achieved acceptable levels of accuracy, precision, and repeatability. However, bending-related nonlinearities, such as hysteresis and dead zones, negatively impacted performance. These outcomes highlight the necessity of implementing closed-loop controllers to mitigate these nonlinearities and enhance the overall system performance.

Future efforts should focus on designing closed-loop control systems to achieve higher accuracy. Additionally, incorporating medical imaging modalities, such as ultrasound imaging, is anticipated to test the biomedical functionalities of the haptic system, enabling comprehensive \textit{in-vitro} and \textit{ex-vivo} experiments. Finally, the quality of teleoperation will be studied by considering factors such as latency, bandwidth, bilateral haptic feedback, stability, robustness, and reliability of the entire system. In particular, a bilateral vision-based force feedback system, \cite{nazari2021image}, for interaction control of the teleoperated system will be of great interest to the authors.

%%%%%%%%%%%%%%%%%%%%%%%%%%%%%%%%%%%%%%%%%%%%%%%%%%%%%%%%%%%%%%%%%%%%%%%%%%%%%%%%%%%%%%%%%%%%%%%%%%
\bmhead{Acknowledgements}

This work was supported in part by the Natural Sciences and Engineering Research Council of Canada (NSERC) under Discovery Grants 2017-06930 and 2019-05562. The authors would like to thank Marcelo Tanglao for his insight and efforts in designing the initial version of the follower device, as well as Taha Abbasi-Hashemi for his insight into the integration of the master and follower within the teleoperation system.

\bmhead{Author Contributions}

\textbf{Ali A. Nazari}: Conceptualization, Methodology, Software, Validation, Formal analysis, Investigation, Data Curation, Writing - original draft, Writing - review \& editing, Visualization, Project Administration. \textbf{Jeremy Catania}: Conceptualization, Methodology, Investigation, Writing - original draft, Writing - review \& editing, Visualization. \textbf{Soroush Sadeghian}: Software, Investigation, Data Curation, Writing - review \& editing. \textbf{Amir Jalali}: Methodology, Validation, Writing - review \& editing. \textbf{Houman Masnavi}: Software, Validation, Formal analysis, Investigation, Writing - original draft, Writing - review \& editing. \textbf{Farrokh Janabi-Sharifi}: Conceptualization, Resources, Writing - review \& editing, Supervision, Funding acquisition. \textbf{Kourosh Zareinia}: Conceptualization, Methodology, Resources, Writing - review \& editing, Supervision, Funding acquisition.

\bmhead{Ethics Statement}
This article does not contain any studies with human participants or animals performed by any of the authors.

\bmhead{Conflict of interest} All authors declare that they have no conflicts of interest.

\bmhead{Permission to Reproduce Material From Other Sources} All figures, images, and content in this paper are original works created by the authors. No material has been reproduced from other sources.

%%%%%%%%%%%%%%%%%%%%%%%%%%%%%%%%%%%%%%%%%%%%%%%%%%%%%%%%%%%%%%%%%%%%%%%%%%%%%%%%%%%%%%%%%%%%%%%%%%

\bibliography{ref}

\end{document}